\documentclass{article}

\PassOptionsToPackage{numbers, compress}{natbib}
\usepackage[final]{nips_2017}

\usepackage[utf8]{inputenc} % allow utf-8 input
\usepackage[T1]{fontenc}    % use 8-bit T1 fonts
\usepackage{hyperref}       % hyperlinks
\usepackage{url}            % simple URL typesetting
\usepackage{booktabs}       % professional-quality tables
\usepackage{amsfonts}       % blackboard math symbols
\usepackage{nicefrac}       % compact symbols for 1/2, etc.
\usepackage{microtype}      % microtypography

\usepackage{microtype,marvosym} 
\usepackage{amsmath,amssymb,graphicx}

\usepackage{tikz,pgfplots}
\pgfplotsset{compat=newest}
\pgfplotsset{plot coordinates/math parser=false}       

\usepackage{natbib}

\definecolor{lred}{RGB}{200,0,0}
\definecolor{dred}{RGB}{130,0,0}
\definecolor{ldre}{RGB}{190.0515  127.5000  127.5000}

\definecolor{dblu}{RGB}{0,0,130}
\definecolor{lblu}{RGB}{127.5000, 127.5000, 192.3975}

\definecolor{dgre}{RGB}{0,130,0} 
\definecolor{dgra}{RGB}{50,50,50}
\definecolor{mgra}{RGB}{100,100,100}
\definecolor{lgra}{RGB}{220,220,220}

\definecolor{MPG}{RGB}{000,125,122}
\definecolor{ora}{HTML}{FF9933}

\definecolor{AMPurple}{HTML}{663366}
\definecolor{Burgundy}{HTML}{993333}
\definecolor{Coffee}{HTML}{7B6049}
\definecolor{ForestGreen}{HTML}{005826}
\definecolor{Lavender}{HTML}{6E6AB1}
\definecolor{PSLightBlue}{HTML}{7DA7D9}

\usepackage{booktabs}

\newcommand{\Exp}{\mathbf{E}}

\newcommand{\cov}{\mathbf{cov}} 
\newcommand{\var}{\mathbf{var}}

\newcommand{\sign}{\operatorname{sign}}

\renewcommand{\Re}{\mathbb{R}}

\newcommand{\diag}{\operatorname{diag}}
\newcommand{\N}{\mathcal{N}}

\newcommand{\B}{\mathcal{B}} 

\renewcommand{\L}{\mathcal{L}}
\newcommand{\D}{\mathcal{D}}
\newcommand{\T}{\mathcal{T}}
\newcommand{\Trans}{^{\intercal}} 
\newcommand{\argmin}{\operatorname*{arg\:min}}

\newcommand{\q}{\quad}

\renewcommand{\vec}{\boldsymbol}

\newcommand{\Id}{\vec{I}}

\renewcommand{\L}{\mathcal{L}}

\usepackage{colonequals}

\newcommand{\ec}{\equalscolon}

\usepackage{nicefrac}

\usetikzlibrary{arrows,shapes,plotmarks}

\tikzset{>=stealth'} 
\tikzstyle{graphnode} = 
   [circle,draw=black,minimum size=22pt,text centered,text
     width=22pt,inner sep=0pt] 
\tikzstyle{var}   =[graphnode,fill=white]
\tikzstyle{obs}   =[graphnode,fill=black,text=white]
\tikzstyle{fac}   =[rectangle,draw=black,fill=black!25,minimum size=5pt]
\tikzstyle{facprior} =[rectangle,draw=black,fill=black,text=white,minimum size=5pt]
\tikzstyle{edge}  =[draw=white,double=black,thick,-]
\tikzstyle{prior} =[rectangle, draw=black, fill=black, minimum size=
5pt, inner sep=0pt]
\tikzstyle{dirprior} = [circle, draw=black, fill=black, minimum
size=5pt, inner sep=0pt]

\DeclareSymbolFont{stmry}{U}{stmry}{m}{n}
\DeclareMathSymbol\leftarrowtriangle\mathrel{stmry}{"5E}
\DeclareMathSymbol\rightarrowtriangle\mathrel{stmry}{"5F}

\renewcommand{\to}{\operatorname*{\rightarrowtriangle}}

\newcounter{PHcomment}

\usepackage{tensor}
\newcommand{\StatexIndent}[1][3]{%
  \setlength\@tempdima{\algorithmicindent}%
  \Statex\hskip\dimexpr#1\@tempdima\relax}

\usepackage{algorithm}
\usepackage{algorithmic}

\usepackage{pifont}
\usepackage{placeins}

\newcommand{\sgd}{{\sc sgd}}
\newcommand{\gd}{{\sc gd}}

\newcommand{\rms}{{\sc RMSprop}}

\newcommand{\eb}{{\sc eb}}

\newlength\figheight
\newlength\figwidth 

\newboolean{PDF} %Deklaration
\newboolean{TIKZEXT} %Deklaration

\setboolean{PDF}{true} % compile pdf figures
\setboolean{TIKZEXT}{false} %Zuweisung
\title{Early Stopping without a Validation Set}

\author{
  Maren~Mahsereci \\
  \texttt{mmahsereci@tue.mpg.de} \\
   Max Planck Institute for Intelligent Systems,\\
   Spemannstra\ss e, T\"ubingen, Germany
   \And
   Lukas~Balles \\
   \texttt{lballes@tue.mpg.de} \\
   Max Planck Institute for Intelligent Systems,\\
   Spemannstra\ss e, T\"ubingen, Germany
   \And
   Christoph~Lassner\thanks{equally affiliated with: Bernstein Center for Computational Neuroscience, Otfried-M\"uller-Str. 25, T\"ubingen, Germany} \\
   \texttt{classner@tue.mpg.de} \\
   Max Planck Institute for Intelligent Systems,\\
   Spemannstra\ss e, T\"ubingen, Germany
   \And
   Philipp Hennig \\
   \texttt{ph@tue.mpg.de} \\
   Max Planck Institute for Intelligent Systems,\\
   Spemannstra\ss e, T\"ubingen, Germany
}

\begin{document}

\maketitle

\begin{abstract}
\emph{Early stopping} is a widely used technique to prevent poor generalization performance when training an over-expressive model by means of gradient-based optimization. 
 To find a good point to halt the optimizer, a common practice is to split the dataset into a training and a smaller validation set to obtain an ongoing estimate of the generalization performance.
  We propose a novel early stopping criterion
  based on fast-to-compute local statistics of the computed gradients and entirely removes the need for a held-out validation set.
  Our experiments show that this is a viable approach in the setting of
  least-squares and logistic regression, as well as neural networks. 
\end{abstract}

\section{Introduction}
\label{sec:introduction}
The training of parametric machine learning models often involves the formal task of minimizing the
expectation of a loss (risk) over a population $p(x)$ of data, of the form
\begin{equation}
  \label{eq:2}
  \mathcal{L}(w) = \Exp_{x\sim p(x)} \left [\ell(w, x)\right],
\end{equation}
where the loss function $\ell(w, x)$ quantifies the performance of parameter vector $w\in\Re^D$ on data point $x$. In practice though, the data distribution $p(x)$ is usually unknown, and Eq.~\ref{eq:2} is approximated by the \emph{empirical risk}:
\begin{equation}
  \label{eq:6}
  L_\D(w) = \frac{1}{M}\sum_{x\in\D} \ell(w, x).
\end{equation}
Here $\D$ denotes a dataset of size $M=\vert\D\vert$ with instances drawn independently from $p(x)$. 
Often there is easy access to the gradient of $\ell$ and gradient-based optimizers can be used to minimize the empirical risk.
The gradient descent (\gd) algorithm, for example, updates an estimate $w_t$ for the minimizer of $L_{\D}$ according to $w_{t+1} = w_t - \alpha_t \nabla L_\D(w_t)$ with 
$\nabla L_\D(w) = \nicefrac{1}{M}\sum_{x\in\D} \nabla \ell(w, x)$,
%\begin{equation}
%  \label{eq:8}
%  \nabla L_\D(w) = \frac{1}{M}\sum_{x\in\D} \nabla \ell(w, x)
%\end{equation}
 and some hand-tuned or adaptive step sizes $\alpha_t$. In practice, however, evaluating $\nabla L_{\D}$ can become expensive for very large $M$ thus making it impossible to make progress in a reasonable time.
Instead, stochastic optimization methods are used, which use coarser but much cheaper gradient estimates by randomly choosing a mini-batch $\B\subset\D$ of size $\vert\B\vert = m \ll M$ from the training set and computing 
$\nabla L_\B(w) = \nicefrac{1}{m}\sum_{x\in\B}\nabla \ell(w, x)$.
%\begin{equation}
%  \label{eq:1}  
%  \nabla L_\B(w) = \frac{1}{m}\sum_{x\in\B}\nabla \ell(w, x).
%\end{equation}
The gradient descent update then becomes $w_{t+1} = w_t - \alpha_t \nabla L_\B(w_t)$ and the corresponding iterative algorithm is commonly known as stochastic gradient descent (\sgd) \citep{robbins1951stochastic}.

\subsection{Overfitting,  Regularization and Early-Stopping}
\label{sec:overf-regul}
Since the risk $\L$ is virtually always unknown, a key question arising when minimizing the empirical risk $L_{\D}$, is how the performance of a model trained on a finite dataset
$\D$ generalizes to unseen data.
Performance can be measured by the loss itself or other quantities, e.g., the mean accuracy in classification problems. 
Typically, to measure the generalization performance a finite \emph{test set} is entirely withheld from the training procedure and the performance of the final model is evaluated on it. 
This test loss, however, is also only an estimator for $\L$ (in the same sense as the train loss) 
%with finite precision and drops in accuracy with decreasing test set size. 
with a finite stochastic error whose variance drops linearly with the test set size.
If the used model is overly expressive, minimizing the empirical risk
(Eq.~\ref{eq:6}) exactly---or close to exactly---will usually result in poor test
performance, since the model overfits to the training data. There is a range of
measures that can be taken to mitigate this effect; textbooks like \citet{bishop2006pattern} give an overview over general concepts, chapter 7 of \citet{Goodfellow2016DeepLearning} gives a comprehensive summary targeted at
deep learning. Some widely used concepts are briefly discussed in the following paragraphs.

\emph{Model selection} techniques choose a model among a hypothesis class which, under some measure, has the closest level of complexity to the given dataset. They alter the form of the loss function $\ell$ in Eq.~\ref{eq:6} over an outer optimization loop (first find a good $\ell$, then optimize $L_{\D}$), such that the final optimization on $L_{\D}$ is conducted on an adequately expressive model.
This can---but does not need to---constrain the number of variables of the model.
In the case of deep neural networks the number of variables can
even significantly exceed the number of training
examples~\citep{alexnet,VGG,googlenet,resnet}.

If the dataset is not sufficiently representative of the data distribution,
an opposite (although not incompatible) approach 
%it is often possible 
is to artificially enrich it to match a complex model.
\emph{Data augmentation} artificially enlarges the training set by
adding transformations/perturbations of the training data. This can range from
injecting noise~\citep{sietsma_noise,denoising_autoencoders} to carefully tuned contrast
and colorspace augmentation~\citep{alexnet}.
%\maren{sometimes this might not be possible if the data generation is not known (medical data?)}

Finally, a widely-used provision against overfitting is to add regularization terms to the objective function that penalize the parameter vector $w$, typically measured by the $l_1$ or $l_2$ norm \citep{weight_decay}.
These terms constrain the magnitude of $w$. They tend to drive individual
parameters toward zero or, in the $l^1$ case, enforce sparsity~\citep{bishop2006pattern,Goodfellow2016DeepLearning}. In linear regression, these concepts are known as least-squares 
and {\sc lasso} regularization~\citep{lasso}, respectively.

Despite these countermeasures, high-capacity models will often overfit in the
course of the optimization process. While the loss on the training set decreases
throughout the optimization procedure, the test loss saturates at some point and
starts to increase again. This undesirable effect is usually countered by
\emph{early stopping} the optimization process, meaning, that for a \emph{given} model,
the optimizer is halted if a user-designed early stopping criterion is met.
This is complementary to the model and data design techniques mentioned above and does not 
undo eventual poor design choices of $\ell$.
It merely ensures that we do not minimize the empirical risk $L_\D$ of a given model beyond the point of best generalization.
In practice, however, it is often more accessible to `early-stop' a high-capacity model for algorithmic purposes or because of restrictions to a specific model class, and thus preferred or even enforced by the model designer.

Arguably the gold-standard of early stopping is to monitor the loss on a \emph{validation set}~\citep{earlystopping_first,pruning_survey,es_butwhen}.
%A widely-used early stopping criterion utilizes the loss of a \emph{validation set}~\cite{earlystopping_first,pruning_survey,es_butwhen} 
For this, a (usually small) portion of the training data is split off and its loss is used as an estimate 
of the generalization loss 
$\L$ (again in the same sense as Eq.~\ref{eq:6}), leaving less effective training data to define the training loss $L_{\D}$.
An ongoing estimate of this generalization performance is then tracked and the optimizer is halted when the generalization performance drops again.
This procedure has many advantages, especially for very large datasets where splitting off a part has minor or no effect on the generalization performance of the learned model.
Nevertheless, there are a few obvious drawbacks. Evaluating the model on the validation set in regular intervals can be computationally expensive.
More importantly, the choice of the \emph{size} of the validation set poses a trade-off:
A small validation set has a large stochastic error, which can lead to a misguided stopping decision.
Enlarging the validation set yields a more reliable estimate of generalization, but reduces the remaining amount of training data, depriving the model of potentially valuable information.
This trade-off is not easily resolved, since it is influenced by properties of the data distribution (the variance $\Lambda$ introduced in Eq.~\ref{eq:9} below) and subject to practical considerations, e.g., redundancy in the dataset.

%Recently \citet{MacDuvAda14} motivated an early-stopping criterion based on the interpretation of (stochastic) gradient descent in the framework of variational inference. 
%It is based on estimating the marginal likelihood by tracking the change in entropy of the posterior distribution of $w$, which is induced by each optimization step.
%It remains unclear, however, if this approach is viable in practice.

Recently \citet{MacDuvAda14} introduced an interpretation of (stochastic) gradient descent in the framework of variational inference. As a side effect, this motivated an early-stopping criterion based on the estimation of the marginal likelihood, which is done by tracking the change in entropy of the posterior distribution of $w$, induced by each optimization step. Since the method requires estimation of the Hessian diagonals, it comes with considerable computational overhead.
%
%It remains unclear, however, if this approach is viable in practice.

The following section motivates and derives a cheap and scalable early stopping criterion which is solely based on local statistics of the computed gradients.
  In particular, it does not require a held-out validation set, thus enabling the optimizer to use \emph{all} available training data.

\section{Model}
\label{sec:model}
\begin{figure}[t]
  \centering
  \includegraphics[scale=1.0]{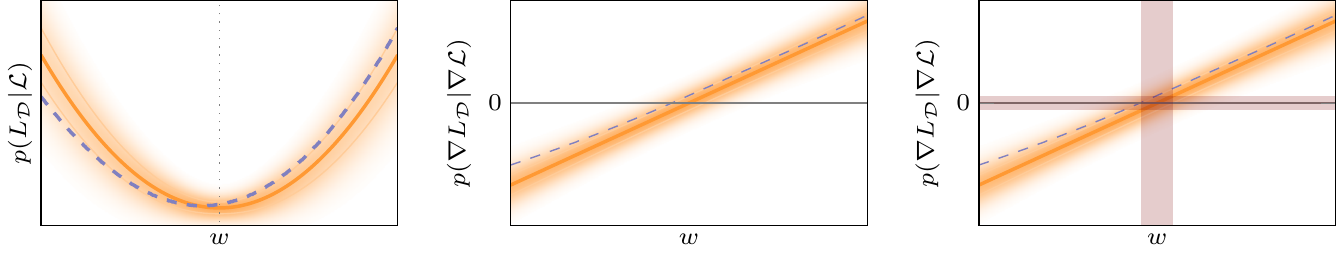}
%\ifthenelse{\boolean{PDF}}
%{
%  \includegraphics[scale=1.0]{pdf_figures/Sketch_trueFunc.pdf}
%%  \includegraphics[scale=1.0]{ext_figs/paper-figure0.pdf}
%}
%{  
% %\tikzset{external/remake next}
%  \setlength{\figwidth}{.99\textwidth}
%  \setlength{\figheight}{.1\textheight}
%  {\scriptsize % 
%    \input{fig/Sketch_trueFunc.tikz}}
%}
  \caption{\emph{Sketch of early stopping criterion}. \textbf{Left:} marginal distribution of function values defined by left expression in Eq.~\ref{eq:9}. Mean $\L$ in thick solid orange, $\pm$1 standard deviations in light orange; pdf as shaded orange. The \emph{full} dataset defines \emph{one} realization of this distribution which is shown in dashed blue (same as $L_{\D}$ of Eq.~\ref{eq:6}). \textbf{Middle:} same as left plot but for corresponding gradients. 
The pdf is defined by the right expression in Eq.~\ref{eq:9} and the corresponding $\nabla L_{\D}$ is shown in dashed blue. 
\textbf{Right:} orange and blue same as middle plot; red shaded ares define desired stopping regions (details in text).
The vertical red shaded area shows the region of $\pm$1 standard deviation of possible minima (where $\nabla L_{\D}$ is likely to be zero). If gradients are within this are, the optimization process is halted. This can be translated into a simple stopping criterion (horizontal shaded area, text for details); if gradients are within this area, the optimizer stops.
}
\label{fig:sketch-plain}
\end{figure}
This section derives a novel criterion for early stopping in stochastic gradient descent. We first introduce notation and model assumptions (\textsection\ref{sec:prer-estim-true}), and motivate the idea of evidence-based stopping (\textsection\ref{sec:when-stop}). Section \ref{sec:gradient-descent} covers the more intuitive case of gradient descent; Section \ref{sec:stoch-grad-desc} extends to stochastic settings.

\subsection{Distribution of Gradient Estimators}
\label{sec:prer-estim-true}
Let $\mathcal{S}$ be some set of instances sampled independently from $p(x)$. The following holds for any $\mathcal{S}$, but specifically for the training set $\D$ or a subsampled mini-batch $\B$ and any validation or test set. Using the same notation as in Eq.~\ref{eq:6}, 
$L_{\mathcal{S}}(w)$ and  $\nabla L_{\mathcal{S}}(w)$
are unbiased estimators of $\L(w)$ and  $\nabla \L(w)$ respectively.
Since the elements in $\mathcal{S}$ are independent draws from $p(x)$, by the Central Limit Theorem
$L_{\mathcal{S}}(w)$ and $\nabla L_{\mathcal{S}}(w)$ are approximately normal distributed according to
\begin{equation}
\begin{split}
  \label{eq:9}
  L_\mathcal{S}(w) & \sim \N\left(\mathcal{L}(w), \frac{\Lambda(w)}{\vert\mathcal{S}\vert}\right)\q\text{and}\q
  \nabla L_\mathcal{S}(w) \sim \N\left(\nabla \mathcal{L}(w), \frac{\Sigma(w)}{\vert \mathcal{S}\vert}\right)
  \end{split}
\end{equation}
with population (co-)variances $\Lambda(w) = \var_{x\sim p(x)} [\ell(w, x)]$ $\in\Re$ and $\Sigma(w) = \cov_{x\sim p(x)}\left[ \nabla \ell(w, x)\right]$ $\in\Re^{D\times D}$, respectively.
%with population (co-)variances $\Lambda(w)$ and $\Sigma(w)$ of function value and gradients respectively
%\begin{equation}
%\begin{split}
%\label{eq:11}
%  \Lambda(w) & = \var_{x\sim p(x)} [\ell(w, x)] \in \Re \q\text{and}\q
%  \Sigma(w) = \cov_{x\sim p(x)}\left[ \nabla \ell(w, x)\right] \in \Re^{D\times D}.
%\end{split}
%\end{equation}
The (co)-variances of $L_{\mathcal{S}}(w)$ and $\nabla L_{\mathcal{S}}(w)$ both scale inversely proportional to the dataset size $|\mathcal{S}|$. 
%, meaning the larger the dataset the more reliable these estimators become.ar
In the population limit $|\mathcal{S}|\to\infty$, Eq.~\ref{eq:9} concentrates on $\L(w)$ and $\nabla \L(w)$.
To simplify notation, the indicator $(w)$ will occasionally be dropped: e.g. $L_{\mathcal{S}}(w) \ec L_{\mathcal{S}}$.

\subsection{When to stop? An Evidence-Based Criterion}
\label{sec:when-stop}

The perhaps obvious but crucial observation at the heart of the criterion proposed below is that even the full, but finite, data-set is just a finite-variance sample from a population: By Eq.~\eqref{eq:9}, the \emph{estimators} $L_{\D}$ and $\nabla L_{\D}$ are approximately Gaussian samples around their expectations $\L$ and $\nabla \L$, respectively.
% Choosing a dataset of size $M:=|\D|$ amounts to sampling one specific realization $[L_{\D}, \nabla L_{\D}]$ from this distribution. 
% A different dataset of same size (but also drawn from $p(x)$) results in a different realization of $[L_{\D}, \nabla L_{\D}]$. 
Figure~\ref{fig:sketch-plain} provides an illustrative, one-dimensional sketch. The
left subplot shows the marginal distribution of function values (Eq.~\ref{eq:9}, left). The true, but usually unknown, optimization objective $\L$
(Eq.~\ref{eq:2}), is the mean of this distribution and is shown in solid
orange. The objective $L_{\D}$ (Eq.~\ref{eq:6}), which is optimized in
practice and is fixed by the training set $\D$, defines \emph{one}
realization out of this distribution and is shown in dashed blue.

In general, the minimizers of  $\L$ and $L_{\D}$ need not be the same. Often, for a finite but large number of parameters $w\in\Re^D$, the loss $L_{\D}$ can be optimized to be very small. When this is the case the model tends to overfits to the training data and thus performs poorly on newly generated (test) data $\T\sim p(x)$ with $\T\cap \D = \varnothing$. A widely used technique to prevent overfitting is to stop the optimization process early. 
%This is complementary to model selection techniques like weight regularization since $\L$ stays unchanged. 
The idea is, that variations of training examples which do not contain information for generalization, are mostly learned at the very end of the optimization process where the weights $w$ are fine-tuned.
In practice the true minimum of $\L$ is unknown, 
however the approximate errors of the estimators $L_{\D}$ and $\nabla L_{\D}$ are accessible at every position $w$. Local estimators for the diagonal of $\Sigma(w)$ have been successfully used before \citep{MahHen2015, 2016arXiv161205086B} and can be computed efficiently even for very high dimensional optimization problems. Here the variance estimator of the gradient distribution is denoted as $\hat{\Sigma}(w)\approx \var_{x\sim p(x)}\left[ \nabla \ell(w, x)\right]$ with
$\hat{\Sigma}(w) = \nicefrac{1}{(|\mathcal{S}|-1)}\sum_{x\in\mathcal{S}} \left(\nabla \ell(w, x) 
- \nabla L_{\mathcal{S}}(w)\right)^{\odot 2}$,
%\begin{equation}
%  \label{eq:3}
%  \hat{\Sigma}(w) = \frac{1}{|\mathcal{S}|-1}\sum_{x\in\mathcal{S}} \left(\nabla \ell(w, x) 
%- \nabla L_{\mathcal{S}}(w)\right)^{\odot 2}
%\end{equation}
where $^{\odot 2}$ denotes the elementwise square and $\mathcal{S}$ is either the full dataset $\D$ or a mini-batch $\B$.

Since the minimizers of $\L$ and $L_{\D}$ are not generally identical, also their gradients will cross zero at different locations $w$. 
The middle plot of Figure~\ref{fig:sketch-plain} illustrates this behavior. Similar to the left plot, it shows a marginal distribution, but this time over gradients (right expression in Eq.~\ref{eq:9}). The true gradient $\nabla \L$ is the mean of this distribution and is shown in solid orange. The \emph{one} realization defined by the dataset $\D$ is shown as dashed blue and corresponds to the dashed blue function values $L_{\D}$ of the left plot. Ideally the optimizer should stop in an area in $w$-space where possible minima are likely to occur, if different datasets of same size were samples from $p$.
In the sketch, this is encoded as the red \emph{vertical} shaded area in the right plot. It is the area around the minimizer of $\L$ where $\nabla\L\pm 1$ standard deviation still encloses zero.

% (red vertical shaded area in right plot of Figure \ref{fig:sketch-plain}). 

Since $\nabla \L$ is unknown however, this criterion is hard to use in practice, and must be turned into a statement about $\nabla L_D$.
Denote the minimizer of $\L$ by $w^*=\argmin_w\L(w)$ and the population variance of gradients at $w^*$ as $\Sigma^*:=\Sigma(w^*)$.
A similar criterion that captures this desiderata in essence is to stop when the collected gradients 
$\nabla L_{\D}$ are becoming consistently very small in comparison to the error $\nicefrac{\Sigma^*}{M}$ 
(red \emph{horizontal} shaded area). Close enough to the minima of $L_{\D}$ and $\L$, the two criteria roughly coincide (intersection of red vertical and horizontal shaded areas). 
A measure for this is the probability
\begin{equation}
  \label{eq:18}
  \begin{split}
    p(\nabla L_{\D}|\nabla\L=0) 
    & = \N\left(\nabla L_{\D}; 0, \frac{\Sigma^*}{M}\right),
  \end{split}
\end{equation}
%\begin{equation}
%  \label{eq:18}
%  \begin{split}
%    p(\nabla L_{\D}) &= \int p\left(\nabla L_{\D}|\nabla \L\right)
%    p\left(\nabla \L\right)\mathrm{d}\nabla \L
%     = \N\left(\nabla L_{\D}; 0, \frac{\Sigma^*}{M}\right).
%  \end{split}
%\end{equation}
of observing $\nabla L_{\D}$, were it generated by a true zero gradient $\nabla \L =0$.
This can be seen as the evidence of the trivial model class $p(\nabla \L)=\delta(\nabla \L)$, 
with $p(\nabla L_{\D}) = \int p\left(\nabla L_{\D}|\nabla \L\right)  p\left(\nabla \L\right)\mathrm{d}\nabla \L$ (in principal more general models can be formulated, which lead to a richer class of stopping criteria).
If gradients  $\nabla L_{\D}$  are becoming too small or, `too probable' (stepping into the horizontal shaded area) the gradients are less likely to still carry information about $\nabla \L$ but rather represent noise due to the finiteness of the dataset, then the optimizer should stop. 
Using these assumptions, the next section derives a stopping criterion for the gradient decent algorithm which then can be extended to stochastic gradient descent as well.

\subsection{Early Stopping Criterion for Gradient Descent }
\label{sec:gradient-descent}
When using gradient descent, the whole dataset is used to compute the gradient $\nabla L_{\D}$ in each iteration. Still this gradient estimator has an error in comparison to the true gradient $\nabla \L$, which is encoded in the covariance matrix $\Sigma$. 
In practice $\Sigma$ is unknown, the variance estimator $\hat{\Sigma}$ described in Section~\ref{sec:when-stop} however is always accessible. 
In addition Eq.~\ref{eq:18} requires the gradient variance $\Sigma^*$ at the true minimum which is unknown in practice. 
Again it can be approximated by $\Sigma(w_t)$ which is the gradient variance at the current position of the optimizer $w_t$. This is a sensible choice if the optimizer is in convergence and already close to a minimum.
Thus, at every position $w$ an approximation to $p(\nabla L_{\D})$ of Eq.~\ref{eq:18} is
\begin{equation}
  \label{eq:7}
  \begin{split}
  p(\nabla L_{\D}(w)) &\approx \prod_{k=1}^D\N\left(\nabla L_{\D}^k(w); 0, \frac{\hat{\Sigma}_{k}(w)}{M}\right).
  \end{split}
\end{equation}
Though being a simplification, this allows for fast and scalable computations since dimensions are treated independent of each other.
To derive an early stopping criterion based only on $\nabla L_{\D}$ we borrow the idea of the previous section that the optimizer should halt when gradients become so small that they are unlikely to still carry information about $\nabla \L$, and combine this with well-known techniques from statistical hypothesis testing. Specifically: stop when 
\begin{equation}
  \label{eq:17}
  \begin{split}
  \log p\left(\nabla L_{\D}\right) 
-\Exp_{\nabla L_{\D}\sim p}\left[\log p\left(\nabla L_{\D}\right) \right] > 0.
  \end{split}
\end{equation}
Here $\Exp[\cdot]$ is the expectation operator. According to Eq.~\ref{eq:17}, the optimizer stops when the logarithmic evidence of the gradients is larger than its expected value, roughly meaning that more gradient samples $\nabla L_{\D}$ lie inside of some expected range. In particular, combining Eq.~\ref{eq:7} with Eq.~\ref{eq:17} and scaling with the dimension $D$ of the objective, gives 
\begin{equation}
  \label{eq:10}
  \begin{split}
  & \frac{2}{D}\left[\log p\left(\nabla L_{\D}\right) 
-\Exp_{\nabla L_{\D}\sim p}\left[\log p\left(\nabla L_{\D}\right) \right]\right]
 = 1-\frac{M}{D}\sum_{k=1}^D\left[\frac{(\nabla L^k_{\D})^2}{\hat{\Sigma}_{k}}\right] 
>0.
  \end{split}
\end{equation}
This criterion (hereafter called \eb-criterion, for `evidence-based') is very intuitive; if all gradient elements lay at exactly one standard deviation distance to zero, then $\sum_{k}\nicefrac{(\nabla L^k_{\D})^2}{\hat{\Sigma}_k}=\sum_{k}\nicefrac{\hat{\Sigma}_k}{M\cdot\hat{\Sigma}_k}=\nicefrac{D}{M}$; thus the left-hand side of Eq.~\ref{eq:10} would become zero and the optimizer would stop. 

We note on the side that Eq.~\ref{eq:10} defines a mean criterion over all elements of the parameter vector $w$. This implicitly assumes that all dimensions converge in roughly the same time scale such that weighing the fractions $f_k:=\nicefrac{M\cdot(\nabla L^k_{\D})^2}{\hat{\Sigma}_{k}}$ equally is justified. If optimization problems deal with parameters that converge at different speeds, like for example different layers of neural networks (or biases and weights inside one layer) it might be appropriate to compute one stopping criterion per subset of parameters which are roughly having similar timescales. 
In Section \ref{sec:multi-layer-perc} we will use this slight variation of Eq.~\ref{eq:10} for experiments on a multi layer perceptron.

\subsection{Stochastic Gradients and Mini-batching}
\label{sec:stoch-grad-desc}
It is straightforward to extend the stopping criterion of Eq.~\ref{eq:10} to stochastic gradient descent (\sgd); the estimator for $\nabla L_{\D}$ is replaced with an even more uncertain $\nabla L_{\B}$ by sub-sampling the training dataset at each iteration.
The local gradient generation is
%\begin{alignat}{2}
%  \label{eq:14}
%  \nabla L_{\D} &= \nabla \L + \epsilon,~
%  &&\epsilon\sim \N\left(0, \frac{\Sigma}{M}\right)\notag\\
%  \nabla L_{\B} &= \nabla L_{\D} + \eta,~
%  &&\eta\sim \N\left(0, \Sigma_{obs}\right)\\
%   &= \nabla \L + \nu,~
%  &&\nu\sim \N\left(0, \frac{\Sigma}{M} +  \Sigma_{obs}\right).\notag
%\end{alignat}
\begin{equation}
  \label{eq:14}
  \begin{split}
  \nabla L_{\B} = \nabla L_{\D} + \eta = \nabla \L + \nu\q\text{with}\q
\eta\sim \N\left(0, \Sigma_{\mathrm{obs}}\right),
\nu\sim \N\left(0, \nicefrac{\Sigma}{M} +  \Sigma_{\mathrm{obs}}\right).
  \end{split}
\end{equation}
Combining this with Eq.~\ref{eq:9} yields $\nicefrac{\Sigma}{M} +  \Sigma_{\mathrm{obs}} = \nicefrac{\Sigma}{m}$. Thus $\Sigma_{\mathrm{obs}}= \frac{M-m}{mM}\Sigma$. Equivalently to Eq.~\ref{eq:18}, \ref{eq:7} and \ref{eq:10}, this results in an early stopping criterion for stochastic gradient descent:
\begin{equation}
  \label{eq:5}
  \begin{split}
  & \frac{2}{D}\left[\log p\left(\nabla L_{\B}\right) 
-\Exp_{\nabla L_{\B}\sim p}\left[\log p\left(\nabla L_{\B}\right) \right]\right]
 = 1-\frac{m}{D}\sum_{k=1}^D\left[\frac{(\nabla L^k_{\B})^2}{\hat{\Sigma}_{k}}\right] 
>0.
  \end{split}
\end{equation}

\emph{Remark on implementation:}
Computing the stopping criterion is straight-forward, given that the variance estimate $\hat{\Sigma}$ is available.
In this case, it amounts to an element-wise division of the squared gradient by the variance, followed by an aggregation over all dimensions.
\citet[\textsection 4.2]{2016arXiv161205086B} comment on this issue and present a solution for computing $\hat{\Sigma}$ in contemporary software frameworks,  that computes the variance estimate implicitly, increasing e.g. the computational cost of a backward pass of a neural network by a factor of about 1.25. 

%\section{Implementation}
%\label{sec:implementation}
%Computing the stopping criterion is straight-forward, given that the variance estimate $\hat{\Sigma}(w)$ is available.
%In this case, it amounts to an element-wise division of the squared gradient by the variance, followed by an aggregation over all dimensions.
%
%The variance estimate can be computed directly using Eq.~\eqref{eq:3} if the gradients $\ell(w, x)$ can be accessed individually for each example $x$.
%In many cases though, most notably in contemporary software frameworks for the training of neural networs, gradients are computed simultaneously and aggregated implicitly for efficiency.
%\cite[\textsection 4.2]{2016arXiv161205086B} comment on this issue and present a solution that computes the variance estimate implicitly, increasing the computational cost of a backward pass by a factor of about 1.25. 

\section{Experiments}
\label{sec:experiments}
For proof of concept experiments, we evaluate the \eb-criterion on a number of standard classification and regression problems.
For illustration and analysis, Sections \ref{sec:linear-least-squares} and  \ref{sec:synthQuadProb} show a least-squares toy problem and large synthetic quadratic problems;
Sections \ref{sec:logistic-regression} and \ref{sec:multi-layer-perc} deal with the more realistic setting of logistic regression on the well-known Wisconsin Breast Cancer Dataset (WDBC) \citep{citeulike:8632439}
%\footnote{\scriptsize{\url{http://archive.ics.uci.edu/ml/datasets/Breast+Cancer+Wisconsin+(Diagnostic)}}} 
and a multi layer perceptron on the handwritten digits dataset MNIST~\citep{lecun1998gradient}.
%\footnote{\scriptsize{\url{http://yann.lecun.com/exdb/mnist/}}}~\citep{lecun1998gradient}.
Section~\ref{sec:sector} contains experiments for logistic regression, as well as for a shallow neural network on the SECTOR dataset \cite{sector}; the SECTOR dataset complements MNIST and WDBC, in the sense, that it has a much less favorable feature-to-datapoint ratio ($\sim 9$); increasing the gains on the generalization performance, when all available training data can be used.

\subsection{Linear Least-Squares as Toy Problem}
\label{sec:linear-least-squares}
\begin{figure}[t]
  \centering
  \includegraphics[scale=1.0]{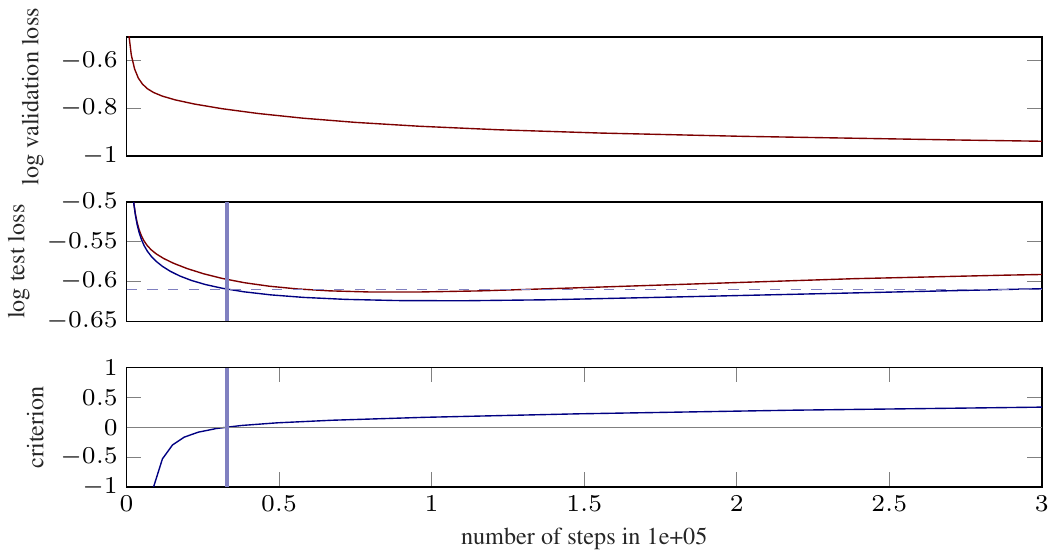}
%\ifthenelse{\boolean{PDF}}
%{
%  \includegraphics[scale=1.0]{pdf_figures/WBCD_LogReg_critAndLoss_GD.pdf}
%%  \includegraphics[scale=1.0]{ext_figs/paper-figure1.pdf}
%}
%{  
%  %\tikzset{external/remake next}
%  \setlength{\figwidth}{.7\textwidth}
%  \setlength{\figheight}{.2\textheight}
%  {\scriptsize % 
%    \input{fig/WBCD_LogReg_critAndLoss_GD.tikz}}
%}
\caption{Results for \emph{logistic regression} on the Wisconsin Breast Cancer dataset. Results for the two variants are color-coded; red for validation set-based early stopping, blue for the evidence-based criterion of Eq.~\ref{eq:10}. The \textbf{middle} plot shows test loss versus the number of optimization steps for both methods.
The  \textbf{top} row shows validation loss; since the validation loss decreases over the whole optimization process it does \emph{not} induce a stopping point. 
The  \textbf{bottom} row shows the evolution of the stopping criterion, inducing a stopping decision indicated by the blue vertical bar.}
\label{fig:logreg}
\end{figure}
\begin{figure}[t]
  \centering
  \includegraphics[scale=1.0]{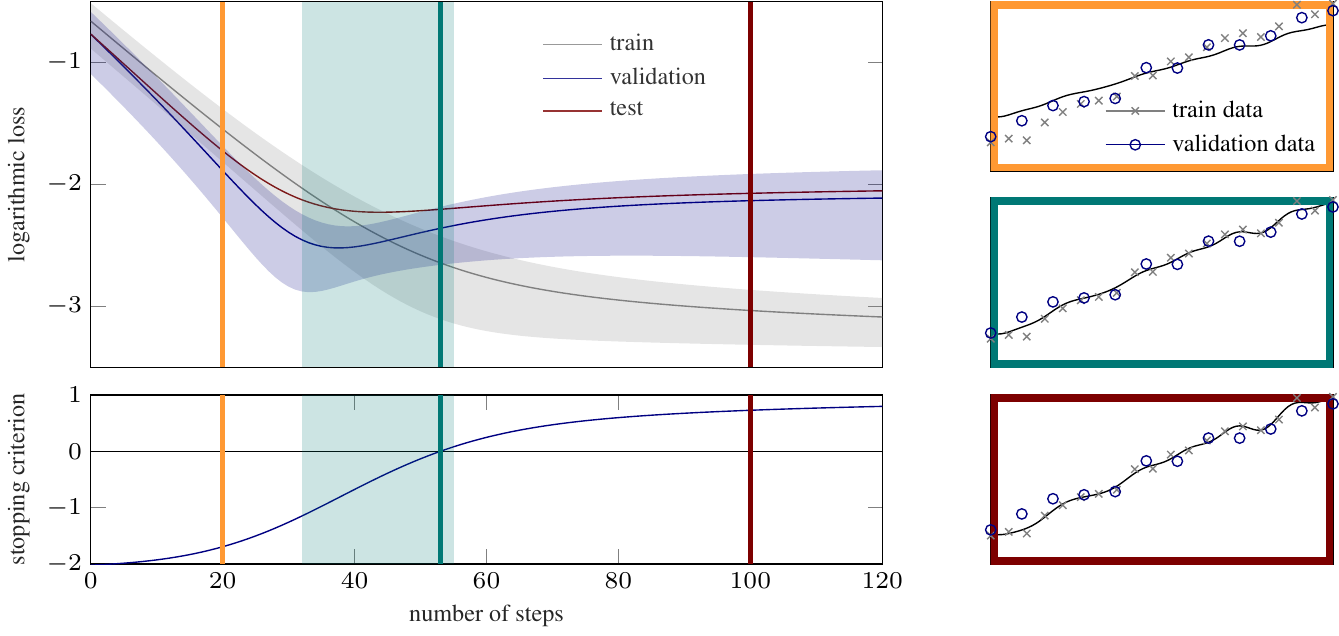}
%\ifthenelse{\boolean{PDF}}
%{
%  \includegraphics[scale=1.0]{pdf_figures/sketch_linRegToyProblem.pdf}
%%  \includegraphics[scale=1.0]{ext_figs/paper-figure2.pdf}
%}
%{  
%  %\tikzset{external/remake next}
%  \setlength{\figwidth}{.95\textwidth}
%  \setlength{\figheight}{.25\textheight}
%  {\scriptsize % 
%   \input{fig/sketch_linRegToyProblem.tikz}}
%}
\caption{\emph{Least-squares toy problem}. \textbf{Top left} logarithmic losses vs. number of optimization steps (colors in legend); shaded areas indicate two standard deviations $\pm 2\sqrt{\nicefrac{\Lambda}{|\mathcal{S}|}}$ of the noise loss estimates computed during the optimization (Eq.~\ref{eq:9}).
%The minimum of the validation loss is suboptimal in terms of the test loss, but falls into an acceptable region (green shaded area).
\textbf{Bottom left:} evolution of the \eb-criterion (Eq.~\ref{eq:10}); green vertical bar indicates the induced stopping point.
%; it is not exactly coinciding with the minimum of the true test loss, but falls into the acceptable region as well.
For the steps marked with color-coded vertical bars, the model fit is illustrated on the \textbf{right column}; orange iteration: sub-optimal fit ($\hat{y}(w)$ in solid dark blue) to the training data (gray crosses); green iteration: fit, when the \eb-criterion of Eq.~\ref{eq:10} indicates stopping; red iteration: the model $\hat{y}$ has already overfitted to the training data.}
\label{fig:linReg-toy-prob}
\end{figure}

%\caption{\emph{Least-squares toy problem:} The \textbf{top left} plot shows logarithmic losses over the number of optimization steps.
%The train loss (gray) decreases throughout the optimization process, whereas the ground-truth test loss (red) saturates after approximately 40 steps and increases thereafter.
%The validation loss (blue) is an uncertain estimate of the test loss; the shaded area indicates two standard deviations $\pm 2\sqrt{\nicefrac{\Lambda}{|\mathcal{S}|}}$ computed during the optimization (Eq.~\ref{eq:9}).
%Notice how the minimum of the validation loss is suboptimal in terms of the test loss, but falls into an acceptable region (green shaded area).
%The \textbf{bottom left} plot shows the evolution of the \eb-criterion (Eq.~\ref{eq:10}) and the green vertical bar indicates the suggested stopping point; it is not exactly coinciding with the minimum of the true test loss, but falls into the acceptable region as well.
%For the steps marked with color-coded vertical bars, we illustrate the model fit on the \textbf{right-hand side}. At the orange iteration the training is still in progress as can be seen from the sub-optimal fit ($\hat{y}(w)$ in solid dark blue) to the training data (gray crosses). The green iteration shows the fit when the \eb-criterion of Eq.~\ref{eq:10} indicates stopping; at the red iteration the model $\hat{y}$ has already overfitted to the training data.}

We begin with a toy regression problem on artificial data generated from a one-dimensional linear function $y$ with additive uniform Gaussian noise.
This simple setup allows us to illustrate the model fit at various stages of the optimziation process and provides us with the true generalization performance, since we can generate large amounts of test data.
We use a largely over-parametrized 50-dimensional linear regression model $\hat{y}(w, x) = w\Trans \phi(x)$ which contain the ground truth features (bias and linear) and additional periodic features with varying frequency. The features $\phi(x) = [1, x, \sin(a_1 x), \cos(a_1 x), \dotsc \sin(a_p(x)), \cos(a_p x)]\Trans$ with $p=24$ obviously define a massively over-parametrized model for the true function and is thus prone to overfitting.
We fit the model by minimizing the squared error, i.e. the loss function is $\ell(w, (x, y)) = \frac{1}{2}(y-\hat{y}(w, x))^2$.
We use 20 samples for training and about 10 for validation, and then train the model using gradient descent.
The results are shown in Figure~\ref{fig:linReg-toy-prob}; both, validation loss, and the \eb-criterion find an acceptable point to stop the optimization procedure, thus preventing overfitting. 

\subsection{Synthetic Large-Scale Quadratic Problem}
\label{sec:synthQuadProb}
\begin{figure}[h]
  \centering
  \includegraphics[scale=1.0]{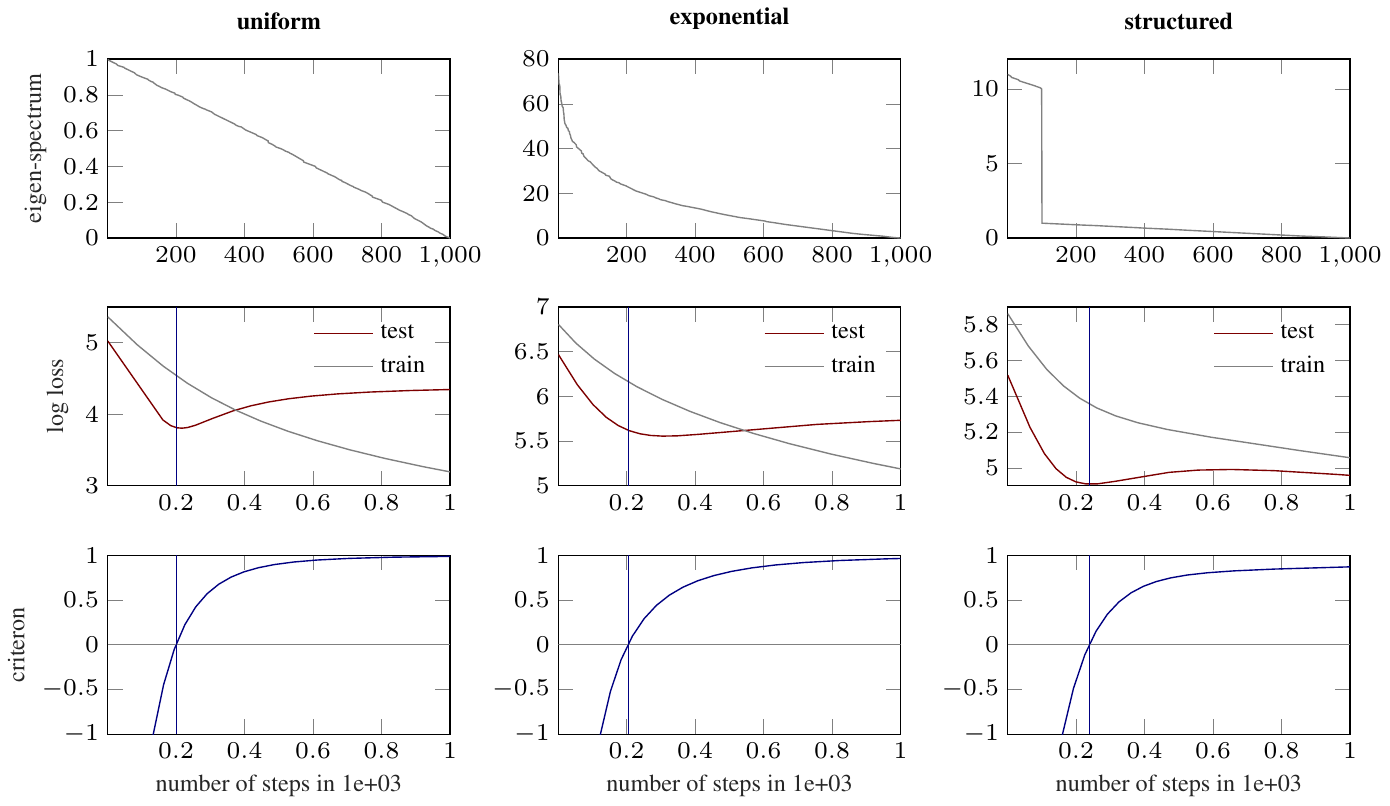}
%\ifthenelse{\boolean{PDF}}
%{
%  \includegraphics[scale=1.0]{pdf_figures/sytheticLargeLinear_uniExpStruc_initFactor0-5.pdf}
%%  \includegraphics[scale=1.0]{ext_figs/paper-figure3.pdf}
%}
%{  
%  %\tikzset{external/remake next}
%  \setlength{\figwidth}{.95\textwidth}
%  \setlength{\figheight}{.3\textheight}
%  {\scriptsize % 
%    \input{fig/sytheticLargeLinear_uniExpStruc_initFactor0.5.tikz}}
%}
\caption{\emph{Synthetic quadratic problem} for three different structures of eigen-spectra: uniform, exponential, structured. \textbf{middle row:} logarithmic (exact) test loss in red and train loss in gray; \textbf{bottom row:} evolution of the \eb-criterion, inducing a stopping decision indicated by the blue vertical bar. 
}
\label{fig:SythLin}
\end{figure}

We construct synthetic quadratic optimization problems of the form
$ \L(w) = \frac{1}{2}(w-w^*)\Trans B(w-w^*)$,
where $B\in\Re^{D\times D}$ is a positive definite matrix and $w^*\in\Re^D$ is the global minimizer of $\L(w)$; the gradient is $\nabla\L = B(w-w^*)$. In this controlled environment we can test the \eb-criterion on different configurations of eigen-spectra, for example uniform, exponential, or structured (a few large, many small eigenvalues); the matrix $B$ is constructed by defining a diagonal matrix $\Gamma\in\Re^{D\times D}$ which contains the eigenvalues on its diagonal, and a random rotation $R\in\Re^{D\times D}$ which is drawn from the Haar-measure on the $D$-dimensional uni-sphere \cite{diaconis1987subgroup}; then $B:=R\Gamma R\Trans$. We artificially define the `empirical' loss $L_\D(w)$ by moving the true minimizer $w^*$ by a Gaussian random variable $\zeta_\D$, such that
$ L_\D(w) = \frac{1}{2}(w-w^*+\zeta_\D)\Trans B(w-w^*+\zeta_\D)\q\text{with}\q \zeta_\D \sim \N\left(0, \Lambda\right)$.
Thus $\nabla L_D = \nabla\L + B\zeta_\D$ is distributed according to $\zeta_\D\sim \N(0, B\Lambda B\Trans)$, and we define $\nicefrac{\hat{\Sigma}}{|D|}:= \diag(B\Lambda B\Trans)$. For experiments we chose $D=10^3$ as input dimension and zero ($w^*=0$) as the true minimizer of $\L$. 
Figure~\ref{fig:SythLin} shows results for three different types of eigen-spectra.
The \eb-criterion performs well across the different type of partially ill-conditioned problems and induced meaningful stopping decisions; this worked well for different noise levels $\Lambda$ (Figure~\ref{fig:SythLin} shows $\Lambda = 10\cdot \Id$; note that the covariance matrix $B\Lambda B\Trans$ of the gradient is dense).

We noticed, however, that another assumption is crucial for the \eb-criterion, which might also explain the slightly early stopping decision for the logistic regressor on WBCD (Figure~\ref{fig:logreg} in subsequent section) and full batch \gd~on MNIST (Figure~\ref{fig:MLP6-BS128}, column 1). Eq.~(\ref{eq:17}) implicitly assumes that (on its path to the minimum of the empirical loss $L_\D$) the optimizer passes by a better minimizer with higher generalization performance; this allows to use variances only (in the form of $\hat{\Sigma}$) in the stopping criterion; there is no information about bias (direction of shift $w^*-w^*_\D$) because this is fundamentally hard to know.

\begin{figure}[h]
  \centering
  \includegraphics[scale=1.0]{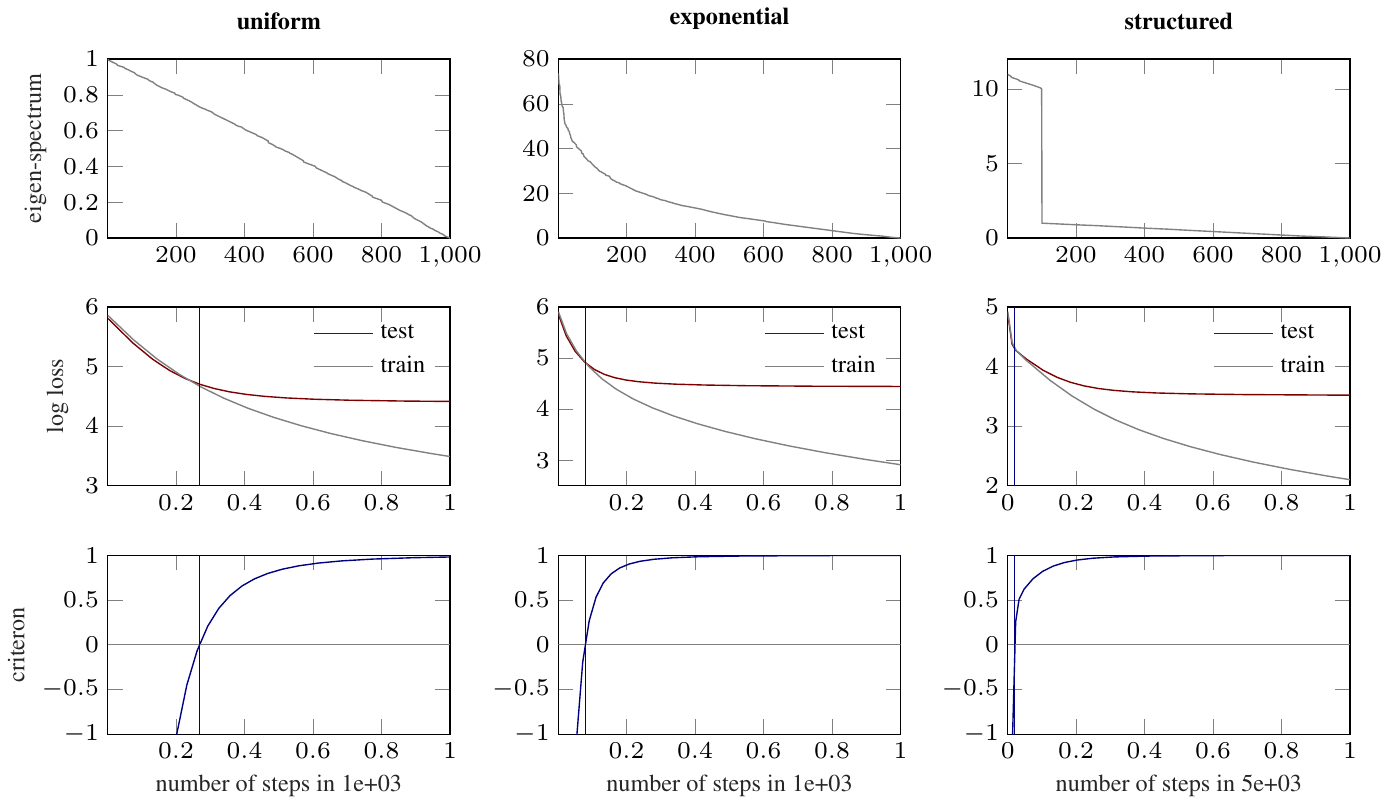}
%\ifthenelse{\boolean{PDF}}
%{
%  \includegraphics[scale=1.0]{pdf_figures/sytheticLargeLinear_uniExpStruc_initFactor5_noOver.pdf}
%%  \includegraphics[scale=1.0]{ext_figs/paper-figure4.pdf}
%}
%{  
%  %\tikzset{external/remake next}
%  \setlength{\figwidth}{.95\textwidth}
%  \setlength{\figheight}{.3\textheight}
%  {\scriptsize % 
%    \input{fig/sytheticLargeLinear_uniExpStruc_initFactor5_noOver.tikz}}
%}
\caption{\emph{Synthetic quadratic problem} for three different structures of eigen-spectra; subplots and colors as in Figure~\ref{fig:SythLin}. Weights are initialized such, that the model can \emph{not} overfit, as can be seen from the exact test loss (red) that flattens out, but does not increase again; the assumptions of the \eb-criterion are violated and it induces a sub-optimal stopping decision.}
\label{fig:SythLin_counter}
\end{figure}

\begin{figure}[h]
  \centering
  \includegraphics[scale=1.0]{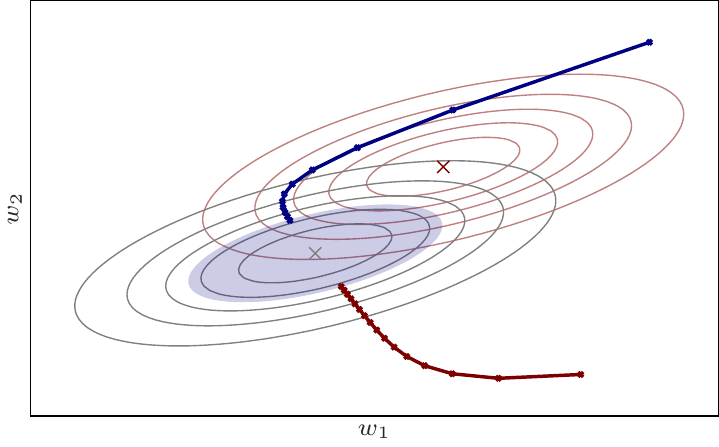}
%\ifthenelse{\boolean{PDF}}
%{
%  \includegraphics[scale=1.0]{pdf_figures/Sketch_2D_assumptions.pdf}
%%  \includegraphics[scale=1.0]{ext_figs/paper-figure5.pdf}
%}
%{  
%  %\tikzset{external/remake next}
%  \setlength{\figwidth}{.5\textwidth}
%  \setlength{\figheight}{.2\textheight}
%  {\scriptsize % 
%    \input{fig/Sketch_2D_assumptions.tikz}}
%}
\caption{Illustration of implicit \emph{early-stopping assumptions:} Contours of the true loss $\L(w)$ in red; contours of the optimizer's objective $L_\D(w)$ in gray; their minimizers $w^*$ and $w_\D^*$ are marked as crosses. The \eb-criterion induces a stopping decision, which is roughly described by the blue shaded area. Blue solid line: path of an optimizer that passes by weights of better generalization performance than $w_\D^*$; it is stopped by the \eb-criterion when it enters the blue shaded area, resulting in better generalization performance. Red solid line: path of an optimizer than can not overfit, since weights were initialized such that $w_\D^*$ yields best generalization performance. The assumptions of the \eb-criterion are violated, and it thus induces a sub-optimal stopping decision that might lead to under-fitting.
}
\label{fig:SythLin_sketch}
\end{figure}

The assumption is usually well justified, primarily because otherwise early stopping would not be a viable concept in the first place; and second because over-fitting is usually associated with `too large' weights (weights are initialized small; and regularizers that pull weights to zero are often a good idea); on the way from small weights (under-fitting) to too large weights (over-fitting), optimizers usually pass a better point with weights of intermediate size.
If the assumption is fundamentally violated the \eb-criterion will stop too early. We can artificially construct this setup by initializing the optimizer with weights that lead to an optimization path that does not lead to \emph{any} over-fitting; this is depicted in Figure~\ref{fig:SythLin_counter}. The setup is identical to the one in Figure~\ref{fig:SythLin} ($B, w^*$ as well as $\zeta_\D$ and $w_\D^*$ are identical); the only difference is the initialization of the weights $w_0$ for the optimization process. Since---with this initialization---the lowest point of $\L$ that can be reached by minimizing $L_\D$ is $w^*_\D$, any early stopping decision will lead to under-fitting. In Figure~\ref{fig:SythLin_counter} the (exact) test loss flattens out and does  not increase again for all three configurations; the assumptions of the \eb-criterion are violated and it induces a sub-optimal stopping decision. Figure~\ref{fig:SythLin_sketch} illustrates these two scenarios in a 2D-sketch.

\subsection{Logistic Regression on WDBC}
\label{sec:logistic-regression}

Next, we apply the \eb-criterion to logistic regression on the Wisconsin Breast Cancer dataset.
The task is to classify cell nuclei (described by features such as radius, area, symmetry, et cetera) as either malignant or benign.
We conduct a second-order polynomial expansion of the original 30 features (i.e., features of the form $x_ix_j$) resulting in 496 effective features.
Of the 569 instances in the dataset, we withhold 369, a relatively large share, for testing purposes in order to get a reliable estimate of the generalization performance.
The remaining 200 instances are available for training the classifier.
We perform two trainining runs: one with early stopping based on a validation set of 60 instances (reducing the training set to 140 instances) and one using the full training set and early stopping with the \eb-criterion derived in Section \ref{sec:gradient-descent}.

If parameters converge at different speeds during the optimization, as indicated in Section \ref{sec:gradient-descent}, it is sensible to compute the criterion separately for different subgroups of parameters.
Generally, if we split the parameters into $N$ disjoint subgroups $S_i\subset \{1, \dotsc D\}$,  and denote $D_i=\vert S_i\vert$, the criterion reads
$\frac{1}{N} \sum_{i=1}^N \left( 1 - \frac{M}{D_i} \sum_{k\in S_i} \left[\frac{(\nabla L^k_{\D})^2}{\hat{\Sigma}_{k}}\right] \right) > 0$.
%\begin{equation}
%\label{eq:999}
%\frac{1}{N} \sum_{i=1}^N \left( 1 - \frac{M}{D_i} \sum_{k\in S_i} \left[\frac{(\nabla L^k_{\D})^2}{\hat{\Sigma}_{k}}\right] \right) > 0.
%\end{equation}
 Since bias and weight gradients usually have different magnitudes they converge at different speeds when trained with the same learning rate. 
For logistic regression, we thus treat the weight vector and the bias parameter of the logistic regressor as separate subgroups. Since the criterion above is noisy we also smooth it with an exponential running average.
The results are depicted in the left-most column of Figure~\ref{fig:MLP6-BS128}.
The effect of the additional training data is clearly visible, resulting in lower test losses throughout the optimization process.
In this scarce data setting the validation loss, computed on a small set of only 60 instances, is clearly misleading (left-most column, top plot).
It decreases throughout the optimization process and, thus, fails to find a suitable stopping point.
The bottom left plot of Fig.~\ref{fig:MLP6-BS128} shows the evolution of the \eb-criterion.
The induced stopping point is not optimal (in that it does not coincide with the point of minimal test loss) but falls into an acceptable region.
Thanks to the additional training data, the test loss at the stopping point is lower than any test loss attainable when withholding a validation set.

\subsection{Multi-Layer Perceptron on MNIST}
\label{sec:multi-layer-perc}

\begin{figure}[t]
  \centering
  \includegraphics[scale=1.0]{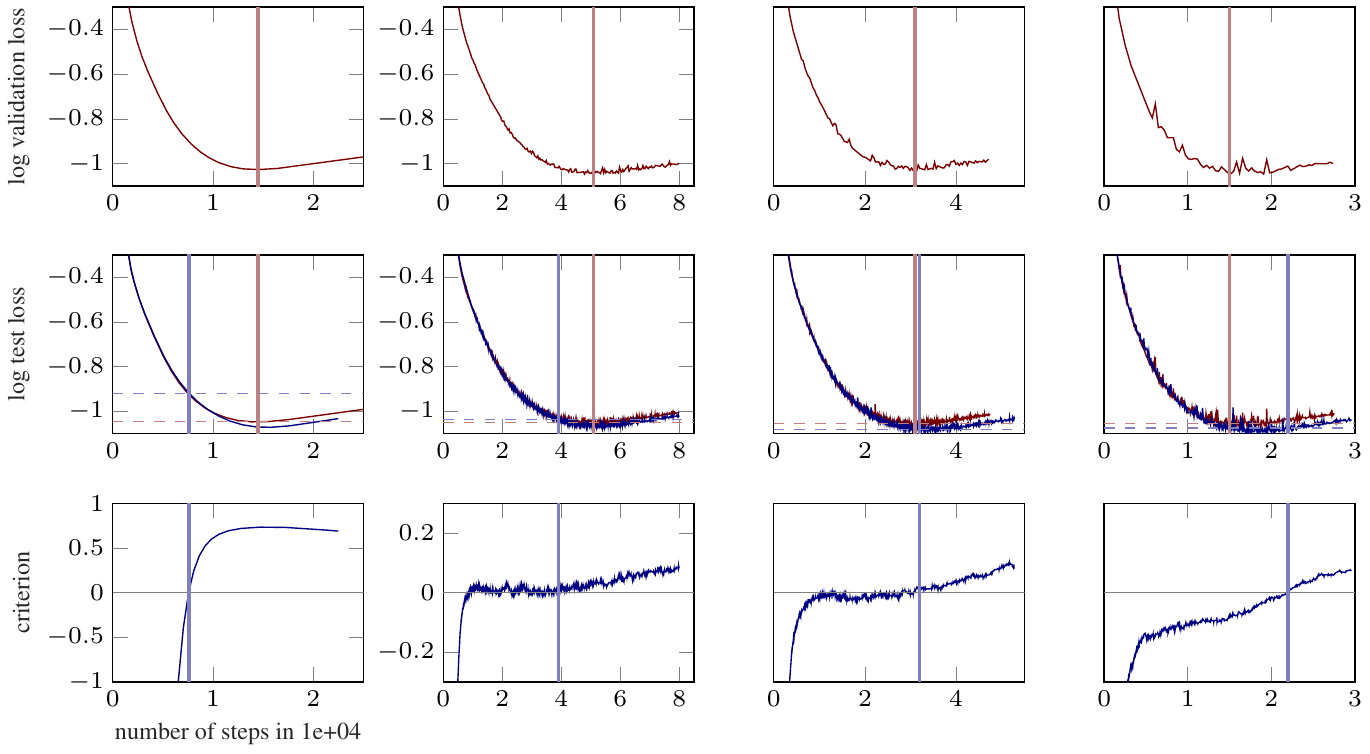}
%\ifthenelse{\boolean{PDF}}
%{
%  \includegraphics[scale=1.0]{pdf_figures/MNIST_MLP6_critAndLoss_GDandSGD.pdf}
%%  \includegraphics[scale=1.0]{ext_figs/paper-figure6.pdf}
%}
%{  
%  %\tikzset{external/remake next}
%  \setlength{\figwidth}{.95\textwidth}
%  \setlength{\figheight}{.3\textheight}
%  {\scriptsize % 
%    \input{fig/MNIST_MLP6_critAndLoss_GDandSGD.tikz}}
%}
\caption{\emph{Multi-layer perceptron} on MNIST: 
\textbf{Column 1:} full batch gradient descent with learning rate 0.01;  
\textbf{columns 2-4} \sgd~with a mini-batch size of 128 and learning rates 0.003, 0.005 and 0.01, respectively. 
Results are color-coded: red for validation set-based early stopping, blue for the \eb-criterion. 
\textbf{Middle row:} logarithmic test loss versus the number of optimization steps for both methods;
\textbf{top row} logarithmic validation loss; minimal point induces a stopping decision (red vertical bar);
\textbf{bottom row:} evolution of the \eb-criterion, stopping decision as blue vertical bar; details in text.
}
\label{fig:MLP6-BS128}
\end{figure}
%\caption{\textbf{Left column:} \emph{Logistic regressor} on WDBC and full batch \gd; 
%\textbf{columns 2-4:} \emph{Multi-layer perceptron} on MNIST and \sgd~with a mini-batch size of 128 and learning rates 0.003, 0.005 and 0.01, respectively. 
%Results are color-coded: red for validation set-based early stopping, blue for the \eb-criterion. 
%\textbf{Middle row:} logarithmic test loss versus the number of optimization steps for both methods;
%\textbf{top row} logarithmic validation loss; minimal point induces a stopping decision (red vertical bar);
%\textbf{bottom row:} evolution of the \eb-criterion, stopping decision as blue vertical bar; details in text.
%Since for MNIST the validation loss is a very good estimator for the generalization loss it very reliably induces good stopping points. Against this strong competitor the \eb-criterion performs as good as or better for the \sgd-runs and slightly worse for the gradient descent run.}
For a non-convex optimization problem, we train a multi-layer perceptron (MLP) on the well-studied problem of hand-written digit classification on the MNIST dataset
%. The inputs to the net are 784-dimensional vectorized gray-scale images of handwritten digits.
($28\times 28$ gray-scale images).
We use a MLP with five hidden layers with 2500, 2000, 1500, 1000 and 500 units, respectively, ReLU activation, and 
a standard cross-entropy loss for the 10 outputs with soft-max activation ($\sim$ 12 million trainable parameters). 
%The ouput layer has 10 units (for the ten digit classes) with soft-max activation and we use standard cross-entropy loss ($\sim$ 12 million trainable parameters). 
We treat each weight matrix and each bias vector of the network as a separate subgroup as described in Section \ref{sec:logistic-regression}.%in the sense of equation \ref{eq:999}.
%The MNIST dataset contains 60k training images
% and a separate test set of 10k images. 
The MNIST dataset contains 60k training images, which we split into 40k-10k-10k for train, test and validation sets.
%For comparison, we perform two training runs for each learning rate: one with early stopping based on a validation set of size 10k (consequently reducing the training set to 40k images) and one using the full training set and early stopping with the \eb-criterion. 
Again, the criterion is smoothed by an exponential running average.

The results for full-batch gradient descent are shown on Column 1 of Figure~\ref{fig:MLP6-BS128}, and \sgd~runs with minibatch size 128 and three different learning rates Column 2-4 of the same Figure.
The relatively large validation set (10k images) yields accurate estimates of the generalization performance. Consequently, the stopping points more or less coincide with the points of minimal test loss.
The reduced training set size leads to only slightly higher test losses.
Since the strength of the \eb-criterion is to utilize the additional training data and the fact, that also validation losses are only inexact guesses of the generalization error, 
both of these points thus favor the early stopping criterion based on the validation loss.
Still, for all three \sgd-runs (columns 2-4 in Figure~\ref{fig:MLP6-BS128}) the \eb-criterion performs as good as or better than the validation set induced method.
An additional observation is that the quality of the stopping points induced by the \eb-criterion varies between the different training configurations.
It is thus arguably not as stable in comparison to setups where the validation loss is \emph{very} reliable. 
%An additional figure for gradient descent runs (full batch training) can be found in the supplements (left out for space constraints).
%
For gradient descent (full training set in each iteration, Column 1 of Figure~\ref{fig:MLP6-BS128}) , the \eb-criterion performs reasonably well, however (an very similarly to the gradient descent runs on the logistic regression on WDBC in Figure~\ref{fig:logreg}) chooses to stop a bit too early, and thus does result in a slightly worse test set performance. 
The difference is not very much (test loss red: $10^{-1.04}$, blue  $10^{-0.92}$) but it also clearly does not outperform the nearly exactly positioned stopping point induced by this well calibrated validation loss.

\subsection{Logistic Regression and Shallow-Net on SECTOR}
\label{sec:sector}
\begin{figure}[h]
  \centering
  \includegraphics[scale=1.0]{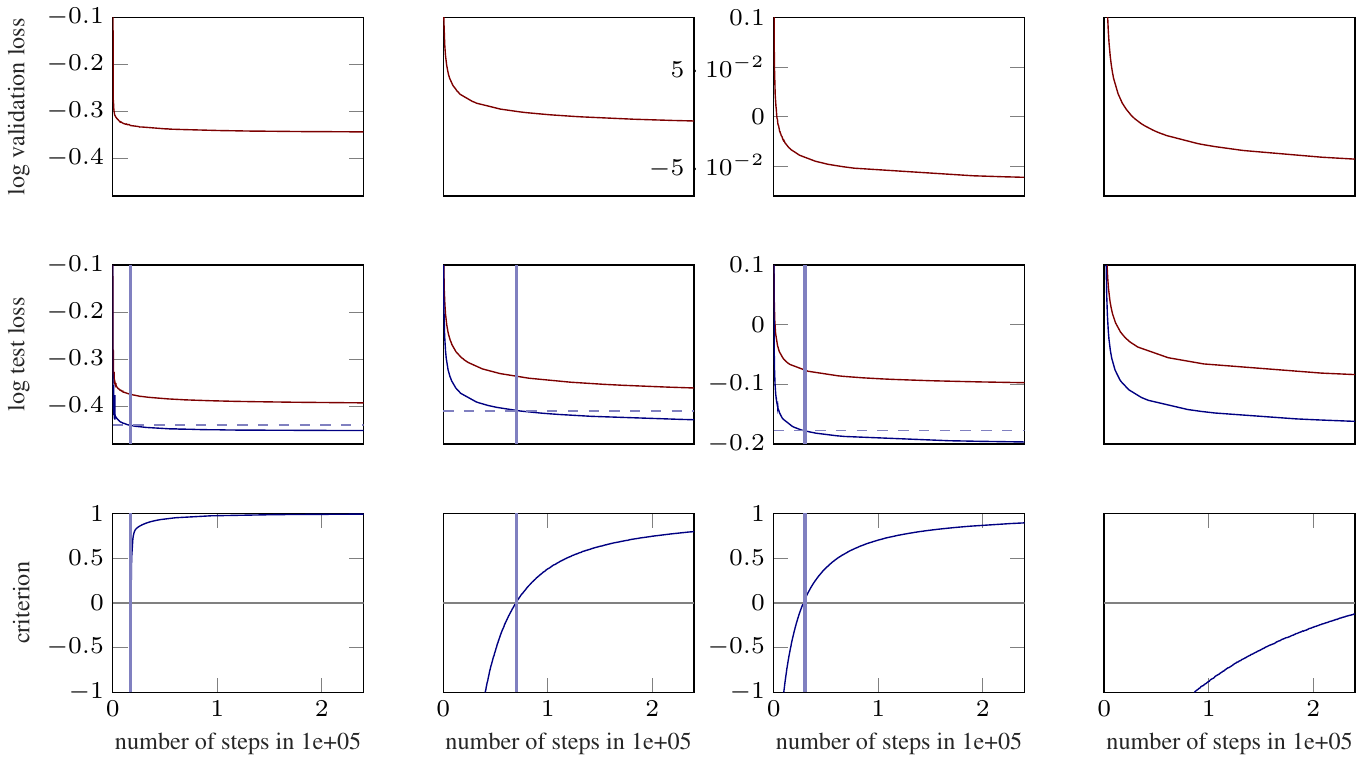}
%\ifthenelse{\boolean{PDF}}
%{
%  \includegraphics[scale=1.0]{pdf_figures/SECTOR_MLP2andLOGREG_critAndLoss_SGD.pdf}
%%  \includegraphics[scale=1.0]{ext_figs/paper-figure7.pdf}
%}
%{  
%  %\tikzset{external/remake next}
%  \setlength{\figwidth}{.95\textwidth}
%  \setlength{\figheight}{.3\textheight}
%  {\scriptsize % 
%    \input{fig/SECTOR_MLP2andLOGREG_critAndLoss_SGD.tikz}}
%}
\caption{\textbf{Colums 1-2:} \emph{Logistic regression} on SECTOR; \sgd~with batch size 128 and learning rates 0.03 and 0.003 respectively;
\textbf{Colums 3-4:} \emph{Shallow net} on SECTOR; \sgd~with batch size 128 and learning rates 0.03 and 0.003 respectively.
Plots and colors as in Figure~\ref{fig:MLP6-BS128}; text for details.
}
\label{fig:sector}
\end{figure}

Finally, we trained a logistic regressor and a shallow fully-connected neural network on the SECTOR dataset\cite{sector}.
%\footnote{{\small \url{https://www.csie.ntu.edu.tw/~cjlin/libsvmtools/datasets/multiclass.html}}}
It contains 6412 training and 3207 test datapoints with 55~197 features each, thus having a less favorable feature-to-datapoint ratio than for example MNIST (784 features vs. 60 000 datapoints). 
The features are extracted from web-pages of companies and the classes describe 105 different industry sectors.
The shallow network has one hidden layer with 200 hidden units; the logistic regressor, thus contains $\sim 5.8$ million, and the shallow net $\sim 11.1$ million trainable parameters.
Experiments are set up in the same style as the ones in Section 3.3 and 3.4. We use $20\%$ of the training data for the validation set; this yields 1282 validation examples and a reduced number of 5130 training examples. Figure~\ref{fig:sector} shows results; columns 1-2 for the logistic regressor and columns 3-4 for the shallow net.
Since the size of the dataset is quite small, the gap between test losses is quite large 
(middle row, full training set (blue), reduced train set, due to validation split (red)). 
Both architectures do not overfit properly, the test loss rather flattens out, although we trained both architectures for very long ($2.5\cdot 10^{5}$ steps) and initialized weights close to zero. 
The \eb-criterion is again a bit too cautious, and induces stopping when the test loss starts to flatten out; but since it allows utilization of \emph{all} training data, it beats the validation set on both architectures.

\subsection{Greedy Element-wise Stopping}
\label{sec:greedy-element-wise}
\begin{figure}[t]
  \centering
  \includegraphics[scale=1.0]{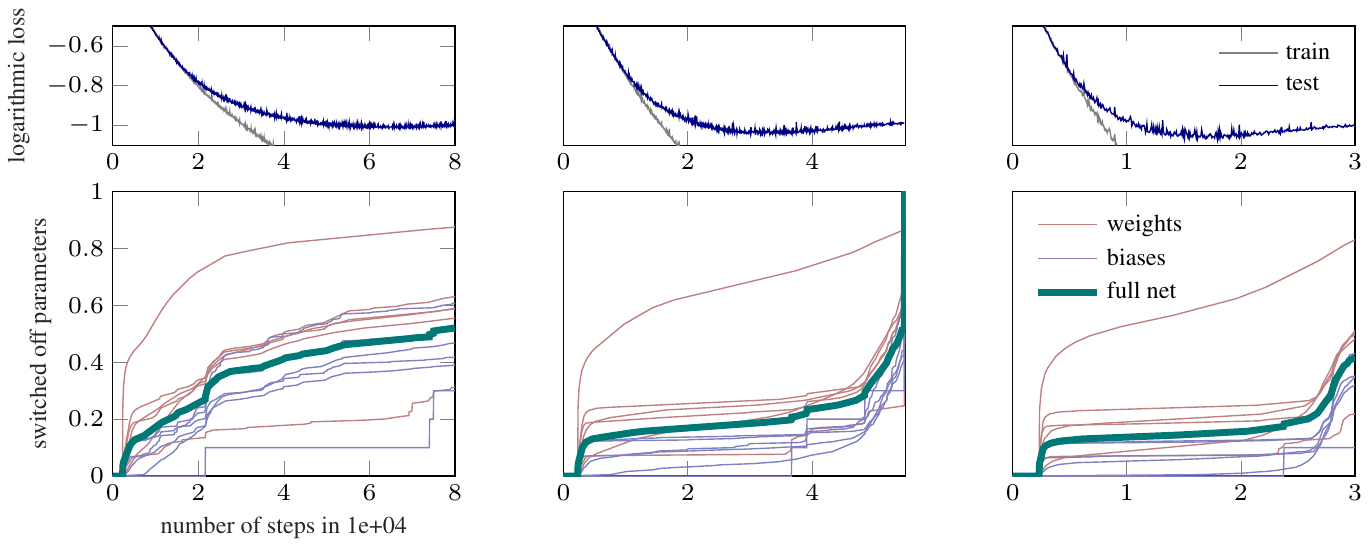}
%\ifthenelse{\boolean{PDF}}
%{
%  \includegraphics[scale=1.0]{pdf_figures/MLP6_critAndLoss_shutOffOpt.pdf}
%%  \includegraphics[scale=1.0]{ext_figs/paper-figure8.pdf}
%}
%{  
%  % \tikzset{external/remake next}
%  \setlength{\figwidth}{.95\textwidth}
%  \setlength{\figheight}{.2\textheight}
%  {\scriptsize % 
%    \input{fig/MLP6_critAndLoss_shutOffOpt.tikz}}
%}
\caption{\emph{Greedy element-wise stopping} for a multi-layer perceptron on MNIST. 
\textbf{Columns:} \sgd~with batch size 128 and learning rates 0.003, 0.005 and 0.01, respectively.
\textbf{Top row} logarithmic training (gray) and test loss (blue).
\textbf{Bottom row} fraction of weights where learning has been shut off by the greedy element-wise stopping; each weight matrix (red), each bias vector (blue), full net (green).
}
\label{fig:MLP6-numOff}
\end{figure}
%\emph{Greedy element-wise stopping:}
For the \eb-criterion, we compute $f_k = m(\nabla L_\B^k)^2 /\hat{\Sigma}_k$ for each gradient element $k$.
This quantity can be understood as a `signal-to-noise ratio' and the \eb-criterion takes the mean over the individual $f_k$.
%
%For the stopping decision, the individual $f_k$ are aggregated (over the whole parameter vector or over subgroups of parameters) and the training is stopped when the \emph{mean} falls below a threshold.
%
As a side experiment, we employ the same idea in an element-wise fashion: we stop the training for an individual parameter $w_k\in\Re$ (not to be confused with the full parameter vector $w_t\in\Re^{D}$ at iteration $t$) as soon as $f_k$ falls below the threshold.
Importantly, 
%To avoid confusion, we want to emphasize that 
this is \emph{not} a sparsification of the parameter vector, since $w_k$ is not set to zero when being switched off but merely fixed at its current value. 
%Since the individual $f_k$ are quite noisy, we smooth them over multiple steps using an exponential moving average; this was not necessary in the aggregate criterion, where variations in the $f_k$ average out.
%The moving averages are initialized at high values, resulting in a warm-up phase where all weights are `active'.
We smooth successive  $f_k$ over multiple steps using an exponential moving average; these averages are initialized at high values, resulting in a warm-up phase where all weights are `active'.
Figure~\ref{fig:MLP6-numOff} presents results;
intriguingly, immediately after the warm-up phase the training of a considerable fraction of all weights (10 percent or more, depending on the training configuration) is being stopped.
This fraction increases further as training progresses. Especially towards the end where overfitting sets in, a clear signal can be seen; the fraction of weights where learning has been stopped suddenly increases at a higher rate.
Despite this reduction in effective model complexity, the network reaches test losses comparable to our training runs without greedy element-wise stopping (test losses in Figure~\ref{fig:MLP6-BS128}).
The fraction of switched-off parameters 
%increases drastically towards the end of the optimization process reaching 
towards the end of the optimization process reaches up to 80 percent in a single layer and around 50 percent for the whole net.

\section{Conclusion}
\label{sec:conclusion}

% Restate the topic.
We presented the \eb-criterion, a novel approach to the problem of determining a good
point for early-stopping in gradient-based optimization.
% Restate the problem.
In contrast to existing methods it does not rely on a held-out validation set
and enables the optimizer to utilize all available training data.
%
% Our solution.
%Instead, 
We exploit fast-to-compute statistics of the observed gradient to assess when 
it
%of an empirical risk 
represents noise originating from the finiteness of the training set, instead of an informative gradient direction.
% Restate the main points.
%To do this, we estimate the evidence of the evaluated gradients under the assumption that the gradient of the true underlying function is zero. 
%
The presented method so far is applicable in gradient descent as well as stochastic
gradient descent settings and adds little overhead in computation, time, and memory consumption.
%The assumptions made are reasonable for most machine learning settings, so that the method is widely applicable.
% Results.
In our experiments, we presented results for linear least-squares fitting,
logistic regression and a multi-layer perceptron, proving the general concept to be viable.
Furthermore, preliminary findings on element-wise early stopping open up the possibility to monitor and control model fitting with a higher level of
detail. 

%\section*{References}

\small

\bibliographystyle{abbrvnat}
\bibliography{bibfile}
\normalsize
\vspace{1cm}
\noindent\rule{\textwidth}{0.4pt}
\begin{center}
%{\huge\textbf{Supplements}}\\
%\vspace{0.1cm}
%{\large---Early Stopping without a Validation Set---}
{\Large\textbf{---Supplements---}}
\end{center}
\section{Comparison to \rms}
\label{sec:conn-diff-rms}
This Section explores the differences and similarities of \sgd+\eb-criterion and \rms. 
This is rather meant as a means for gaining a better intuition, and not for comparing them as competitors; both methods were derived for different purposes and could be combined in principle.

\subsection{Non-Greedy Elementwise \eb-Criterion}
\label{sec:non-greedy-elem}
The  \emph{non-greedy} elementwise \eb-criterion can be formulated as
\begin{equation}
  \label{eq:15}
  \begin{split}
  c_{t}  & = \beta c_{t-1} + (1-\beta)\left(1-f^{\text{\eb-crit}}_t \right)\\
  w_{t+1} & = w_{t} - \alpha \cdot \mathbb{I}\left[c_t\leq 0\right]\odot \nabla L_{\B} (w_t)
  \end{split}
\end{equation}
for some conservative smoothing constant $\beta\in(0, 1)$, usually $\beta\approx 0.999$, or $0.99$, learning rate $\alpha$, 
and the fraction $f^{\text{\eb-crit}}_t := |\B|[\nabla L_{\B}(w_t)^{\odot 2}\oslash \hat{\Sigma}(w_t)]$ as defined in Section~\ref{sec:greedy-element-wise}.
The symbol `$\oslash$' denotes elementwise division and $\mathbb{I}[\cdot]$ is the indicator function. 
In contrast to the greedy implementation of Section~\ref{sec:greedy-element-wise}, where switched-off learning rates stayed switches off, Eq.~\ref{eq:15} allows learning to be switched on again.

\subsection{Learning Rate Damping in \rms}
\label{sec:learn-rate-damp}
\rms~\cite{rmsprop}is a well known optimization algorithm that scales learning rates elementwise by an exponential running average of gradient magnitudes; specifically:
\begin{equation}
  \label{eq:4}
  \begin{split}
  v_{t}  & = \gamma v_{t-1} + (1-\gamma)\nabla L_{\B}(w_t)^{\odot 2}\\
  w_{t+1} & = w_{t} - \alpha \nabla L_{\B}(w_t) \oslash \sqrt{v_t},
  \end{split}
\end{equation}
again for some smoothing constant $\gamma\in(0, 1)$, usually $\gamma\approx 0.95$, and learning rate $\alpha$. Let $z^{\mathrm{max}}_t$ be the largest element of the factor $z_t:=1\oslash \sqrt{v_t}$, then the second line of Eq.~\ref{eq:4} can be rewritten  as
\begin{equation}
  \label{eq:12}
  \begin{split}
  w_{t+1} & = w_{t} - \alpha z_t^{\mathrm{max}}\left(\frac{z_t}{z_t^{\mathrm{max}}}\right)\odot \nabla L_{\B}(w_t).
  \end{split}
\end{equation}
The fraction $f^{\text{\rms}}_t:=\left(\nicefrac{z_t}{z_t^{\mathrm{max}}}\right)\in(0, 1]$ describes the scaling of learning rates relative to the largest one:   
if the $i^{\text{th}}$ element of $f^{\text{\rms}}_t$ is very small, the learning of the corresponding parameter is damped heavily relative to a 
full step of size $\alpha z_t^{\text{max}}$. This can be interpreted as `switching-off' the learning of these parameters,
%of parameters whose fractions $f^{\text{\rms}}_t$ are very small, 
similarly to the elementwise \eb-criterion. 

\subsection{Connections  and Differences}
\label{sec:conn-diff}
The following table gives a rough overview over the possible set of learning rates for each method.
 \begin{center}
   \renewcommand{\arraystretch}{1.5}
    \begin{tabular}{p{2.4cm} p{2.8cm} p{2.8cm} p{3.7cm}}
%      \textbf{method} & \textbf{step size domain}& \textbf{maximal step size} & \textbf{minimal step size}\\
      method & step size domain& maximal step size & minimal step size\\
      \hline
      \sgd          & $\{\alpha\}$                   & $\alpha$                 & $\alpha$ \\
      \sgd+\eb-crit & $\{0,\alpha\}$                & $\alpha$                 & $0$ ~~(only when converged)\\
      \rms          & $(0, \alpha z_t^{\text{max}}]$ & $\alpha z_t^{\text{max}} $ & $>0$ 
    \end{tabular}
\end{center}
The table shows, that \sgd+\eb-criterion is a very minor variation of \sgd, in the sense that it can also set the learning rate to zero, but only for \emph{converged} parameters to prevent overfitting. It does not improve the convergence properties of \sgd~while it is still training, since the sizes of the `active' learning rates remain unchanged.
Specifically, it does not explicitly encode curvature, or other geometric properties of the loss. 

In contrast to this, \rms~also adapts the \emph{absolute value} of the largest possible step at every iteration by a varying factor $z_t^{\text{max}}$, and scales the other steps relative to it.
It is based on the steepest descent direction in $w$-space, measured by a weighted norm, where the weight matrix is the inverse Fisher information matrix $F_t$ at ever position $w_t$.\footnote{If the loss $\ell$ can be interpreted as negative log likelihood, this is an approximation to the steepest descent direction in the distribution space, where an approximation to the KL-divergence defines a measure.} If the learned conditional distribution approximates the true conditional data-distribution well, $F_t$ also approximates the expected Hessian of the loss  \cite{DBLP:journals/corr/Martens14}. \rms~thus encodes geometric information, which allows for faster convergence compared to \sgd.

Another interpretation of \rms, which in spirit is much closer to the \eb-criterion, has recently been formulated by \citet{2017arXiv170507774}.
It is possible to associate the \rms-update of Eq.~\ref{eq:4} with local gradient and variance estimators, according to
\begin{equation}
  \label{eq:19}
  - \alpha \nabla L_{\B}(w_t) \oslash \sqrt{v_t} \approx  - \alpha\frac{\sign[\nabla \L(w_t)]} {\sqrt{1 + \diag[\Sigma(w_t)]\oslash |\B|\nabla\L(w_t)^{\odot 2}}} 
\end{equation}
since 
\begin{equation}
  \label{eq:20}
  \begin{split}
   \nabla L_{\B}(w_t)&\approx\Exp_{x\sim p(x)} \left [\nabla \L_{\B}(w_t)\right]=\nabla\L(w_t),\q\text{and}\\
   v_t&\approx \Exp_{x\sim p(x)} \left [ \nabla L_{\B}(w_t)^{\odot 2}\right] =  \nabla \L(w_t)^{\odot 2} + \frac{\diag[\Sigma(w_t)]}{|\B|} .  
  \end{split}
\end{equation}
The fraction on the right hand side of Eq.~\ref{eq:19} contains the term $\nicefrac{1}{\text{snr}_t}:=\diag[\Sigma(w_t)]\oslash |\B|\nabla\L(w_t)^{\odot 2}$, which closely resembles the inverse of $f_t^{\text{\eb-crit}}$. Thus gradients with a small signal-to-noise ratio $\text{snr}_t$ get shortened; noise free gradients induce steps of equal(!) size $-\alpha\cdot\sign[\nabla \L(w_t)]$ in every direction (note, that they are independent of the magnitude of $\nabla L_{\B}$); \rms~thus can be seen as elementwise stochastic gradient-sign estimators, which are mildly damped if noisy.

We have now explored algebraic, as well as behavioral connections between \sgd+\eb-criterion and \rms;
the following paragraph summarizes the above points and lists some noteworthy distinctions: 

\emph{Geometry encoding:}
\rms~ encodes geometric information about the objective and can be loosely associated with second order methods that perform an approximate diagonal preconditioning at every iteration.
Alternatively it can be interpreted as stochastic sign estimator, scaling each step with the inverse gradient magnitude, and damping due to noise.
In contrast to this, the \eb-criterion is just a mild add-on to \sgd; it does not alter learning rates due to curvature or other geometric effects.

\emph{Mild damping vs. stopping:} 
The \eb-criterion defines a strict threshold, justified by a statistical test, when learning should be terminated.
\rms~defines a vaguer version, in the sense, that the optimizer should move somewhat `less' into directions of uncertain gradients. 
Even if the signal-to-noise ratio $\text{snr}_t$ falls well below the threshold of the stopping decision induces by the \eb-criterion (roughly $\text{snr}_t < 1$), \rms~just reduces the step proportional to the inverse if the square root $\sim (1+\nicefrac{1}{\text{snr}}_t)^{-\nicefrac{1}{2}}$ (e.g. for $\text{snr}_t = 0.5$ (\eb-crit stops), the \rms-step gets reduced by a factor of only $\nicefrac{1}{\sqrt{3}}\approx 0.6$). 
%Thus, even though \rms~shortens high variance directions, they do not get damped enough to prevent overfitting the objective to the data. 

\emph{Smoothing and bias:}
The derivation of Eq.~\ref{eq:19} omits the geometric smoothing contribution of $\gamma$ which is present in the \rms-update in Eq.~\ref{eq:4}.
In contrast to this, the \eb-criterion relies on local (non-smoothed) computations of $\hat{\Sigma}(w_t)$; this is essential to a stopping decision, since large gradient-samples are usually associated with large variances as well. 
Smoothing the latter would thus bias learning towards following large gradients; in case of \rms~it does bias towards larger steps for high variance samples.

The views presented above, give insight on the internal workings of \rms~as well as the \eb-criterion.
%but it remains unclear, how much of \rms's performance gain over \sgd~is due to encoding of geometric effects, or due to mild cautiousness in noisy parameter directions. 
It is apparent, that, even though \rms~shortens high variance directions, they do not get damped enough to prevent overfitting the objective to the data. 
%The \eb-criterion defines a strict, and well defined overfitting decision, but does not alter the underlying optimizer during training.

\subsection{Empirical Comparison}
\label{sec:empirical-comparison}
\begin{figure}[h]
  \centering
  \includegraphics[scale=1.0]{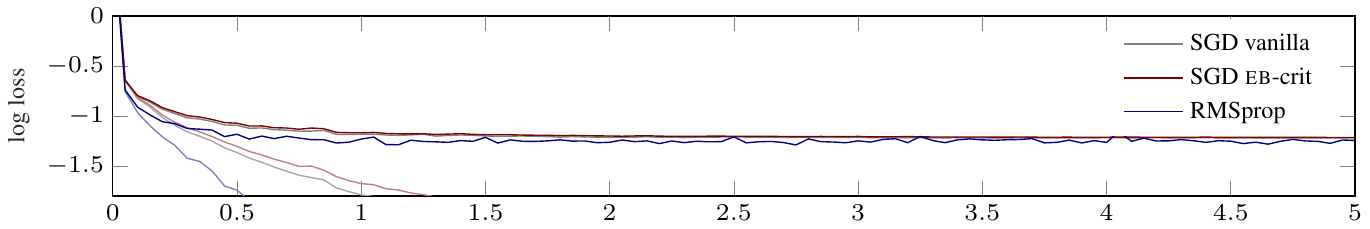}\\%
  \includegraphics[scale=1.0]{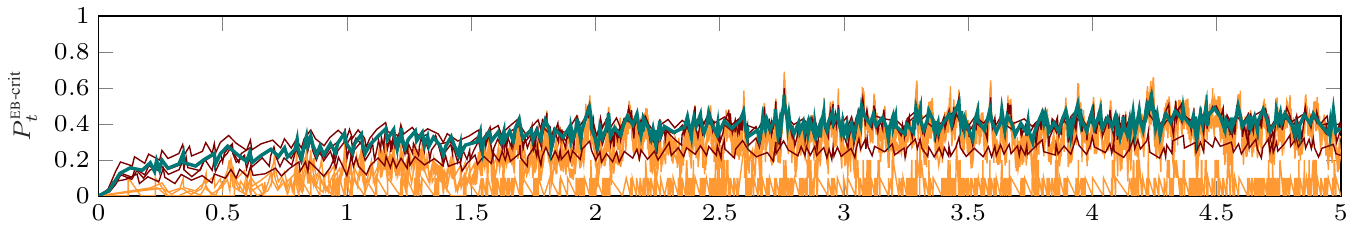}\\%
  \includegraphics[scale=1.0]{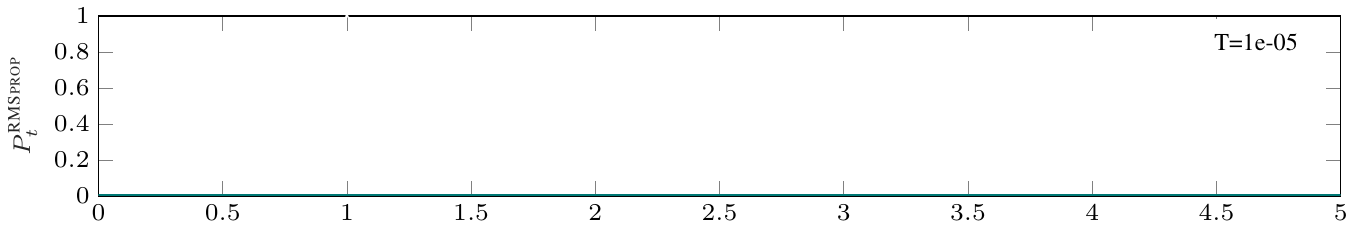}\\%
  \includegraphics[scale=1.0]{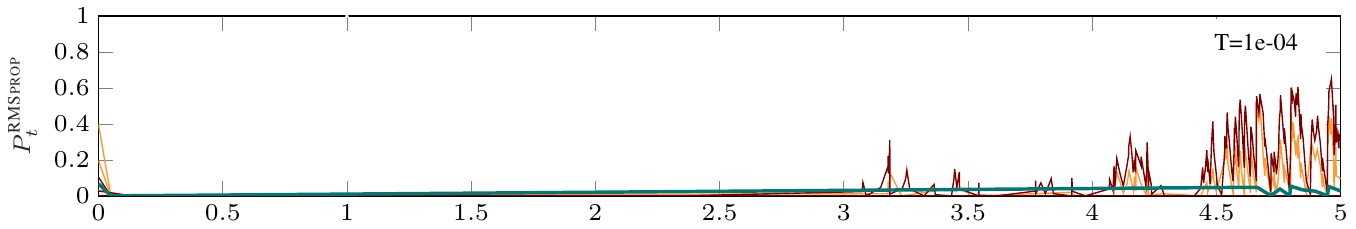}\\%
  \includegraphics[scale=1.0]{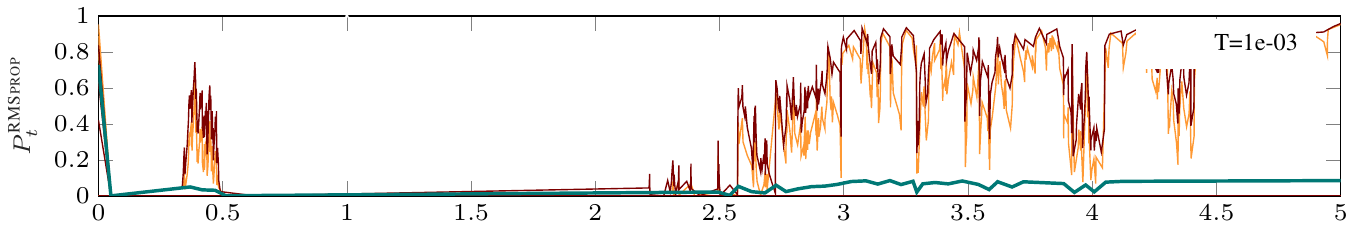}\\%
  \includegraphics[scale=1.0]{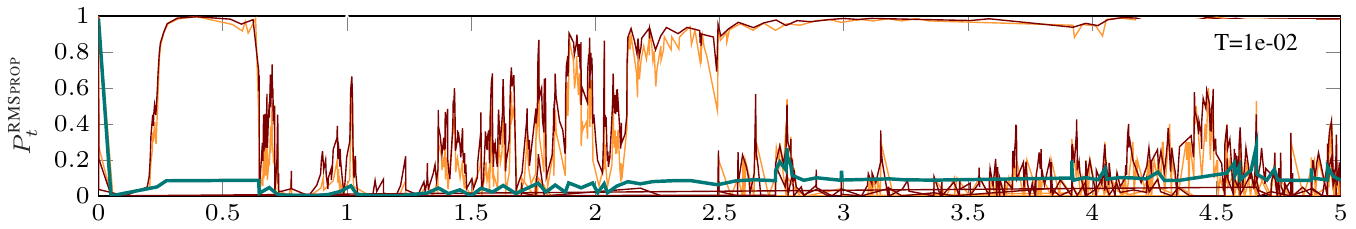}\\%
  \includegraphics[scale=1.0]{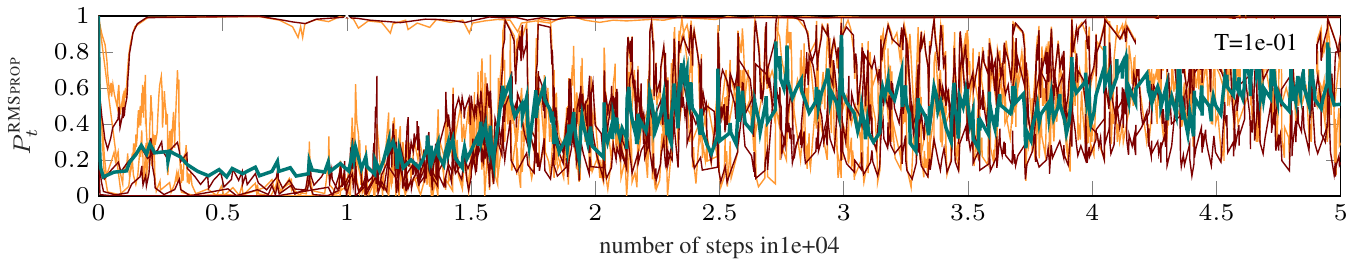}%
%\ifthenelse{\boolean{PDF}}
%{
%  \includegraphics[scale=1.0]{pdf_figures/compare_RMSprop_EBcrit_gamma099_1.pdf}\\%
%  \includegraphics[scale=1.0]{pdf_figures/compare_RMSprop_EBcrit_gamma099_2.pdf}\\%
%  \includegraphics[scale=1.0]{pdf_figures/compare_RMSprop_EBcrit_gamma099_3.pdf}\\%
%  \includegraphics[scale=1.0]{pdf_figures/compare_RMSprop_EBcrit_gamma099_4.pdf}\\%
%  \includegraphics[scale=1.0]{pdf_figures/compare_RMSprop_EBcrit_gamma099_5.pdf}\\%
%  \includegraphics[scale=1.0]{pdf_figures/compare_RMSprop_EBcrit_gamma099_6.pdf}\\%
%  \includegraphics[scale=1.0]{pdf_figures/compare_RMSprop_EBcrit_gamma099_7.pdf}%
%}
%{  
%  %\tikzset{external/remake next}
%  \setlength{\figwidth}{.95\textwidth}
%  \setlength{\figheight}{.08\textheight}
%  {\scriptsize % 
%    \input{fig/compare_RMSprop_EBcrit_gamma0.99_1.tikz}\\%
%    \input{fig/compare_RMSprop_EBcrit_gamma0.99_2.tikz}\\%
%    \input{fig/compare_RMSprop_EBcrit_gamma0.99_3.tikz}\\%
%    \input{fig/compare_RMSprop_EBcrit_gamma0.99_4.tikz}\\%
%    \input{fig/compare_RMSprop_EBcrit_gamma0.99_5.tikz}\\%
%    \input{fig/compare_RMSprop_EBcrit_gamma0.99_6.tikz}\\%
%    \input{fig/compare_RMSprop_EBcrit_gamma0.99_7.tikz}%
%}
%}
\caption{\emph{Comparison of \rms~and \sgd+\eb-criterion} on a multi-layer perceptron on MNIST; batch size is 120.
\textbf{Top row:} logarithmic training loss (light colors) and test loss (corresponding dark colors) for vanilla \sgd~(gray), \sgd+\eb-criterion (red) and \rms(blue).
\textbf{Row 2:} fraction of weights $P_t^{\text{\eb-crit}}$ where learning has been shut off by the elementwise stopping; each weight matrix (red), each bias vector (blue), full net (green). 
\textbf{Row 3-7:} same as row 2, but for $P_t^{\text{\rms}}$ for different choices of $T$ (see legend).
}
\label{fig:compareRMS}
\end{figure}

For an empirical comparison, we run \rms, \sgd~with elementwise \eb-criterion (as in Eq.~\ref{eq:15}), and an instance of vanilla \sgd~on a multi-layer-perception on MNIST, similar to the setup in Section \ref{sec:multi-layer-perc}. 
For the \sgd~instance that uses the \eb-criterion, the fraction of switched-off parameters is defined as
\begin{equation}
  \label{eq:16}
  P_t^{\text{\eb-crit}} := \frac{1}{D}\sum_{i=1}^D\mathbb{I}\left[c_{i, t}\leq 0\right].
\end{equation}
%We track the fraction $f^{\text{\rms}}_t$. 
The percentage of `switched-off' parameters for \rms~can be roughly described as the fraction $P_t^{\text{\rms}}$ of parameters, whose $f^{\text{\rms}}_t$ (defined in Section~\ref{sec:learn-rate-damp}) lie below a threshold $T\in(0, 1)$
\begin{equation}
  \label{eq:13}
  P_t^{\text{\rms}} := \frac{1}{D}\sum_{i=1}^D\mathbb{I}\left[f^{\text{\rms}}_{i, t}<T\right].
\end{equation}
The same smoothing factor $\gamma=\beta=0.99$ was used for both methods, for a meaningful comparison.
Figure~\ref{fig:compareRMS} depicts results; the first row shows training losses (light colors) and test losses (corresponding dark colors) of all three methods.
Rows 3-7 show the evolution of $P_t^{\text{\rms}}$ for five choices of $T=[10^{-1}, 10^{-2}, 10^{-3}, 10^{-4}, 10^{-5}]$; the second row shows $P_t^{\text{\eb-crit}}$. As mentioned above, in contrast to the `greedy' implementation of Section~\ref{sec:greedy-element-wise} (switched-off learning rates, stayed switched-off), and for a more natural comparison to \rms, we allowed learning rates to be switched \emph{on} again as well.
The results for $P_t^{\text{\rms}}$ and $P_t^{\text{\eb-crit}}$ are color coded as in Figure~\ref{fig:MLP6-numOff} of the main paper: green for the full net, and additionally red for weight matrices and orange for biases per layer.

The test losses of vanilla \sgd~and \sgd+\eb-criterion are almost identical, while the training loss of \sgd+\eb-criterion is a bit more conservative than the one of vanilla \sgd; this is expected, since the \eb-criterion ideally should not impair generalization performance, but might lead to larger training losses at convergence, due to the overfitting prevention.
Already at the beginning of the training \sgd+\eb-criterion switches off about 10-20\% of all learning rates; after that, the fraction increases to about 50\% (green line, second row); since the \eb-criterion only detects convergence, the curve is quite monotonic, exhibiting not significant jumps.

\rms~converges a bit faster, as it is expected. Also the plots for $P_t^{\text{\rms}}$ are richer in structure. Especially one layer seems to have significantly smaller learning rates for both, biases and weights, than the other layers. Overall the difference between the largest learning rate and all others tends to roughly increase over the optimization process (especially for $T=10^{-1}$, green line, last row). There are also significant jumps in all the curves, in contrast to the rather monotonic increasing line of \sgd+\eb-criterion. This indicates nontrivial scaling of the absolute, as well as relative sizes of learning rates throughout the optimization process;
also, no learning rate is smaller than $10^{-5}$ times the largest one at each iteration (third row, green line at exactly zero).

In the future a combination of both---learning rate scaling and overfitting prevention---i.e. combining the \eb-criterion with advanced search direction like \rms, is desirable.
%Explore connections further and combine both approaches.

\end{document}